%% file: main.tex
\documentclass{article}

     \PassOptionsToPackage{numbers, compress}{natbib}
\usepackage[preprint]{neurips_2026}

\usepackage[utf8]{inputenc} % allow utf-8 input
\usepackage[T1]{fontenc}    % use 8-bit T1 fonts

\usepackage{url}            % simple URL typesetting
\usepackage{booktabs}       % professional-quality tables
\usepackage{amsfonts}       % blackboard math symbols
\usepackage{nicefrac}       % compact symbols for 1/2, etc.
\usepackage{microtype}      % microtypography
\usepackage{xcolor}         % colors

\usepackage{amsmath}
\usepackage{graphicx}
\usepackage{wrapfig}
\usepackage{mathtools}
\usepackage{amsthm}
\usepackage{changepage}
\usepackage{algorithm}
\usepackage{algorithmic}
\usepackage{amssymb}
\usepackage{array}
\usepackage{bm}
\usepackage{hyperref}       % hyperlinks
\usepackage{cleveref}
\usepackage{subcaption}
\usepackage{bbm}
\usepackage{dsfont}
\input{tex/macro}

\definecolor{BrickRed}{rgb}{0.6,0,0}
\definecolor{RoyalBlue}{rgb}{0,0,0.8}
\hypersetup{colorlinks, citecolor=RoyalBlue, linkcolor=RoyalBlue}

\title{Interaction-Aware Influence Functions\\for Group Attribution}

\author{%
  Jaeseung Heo$^{1}$
  \quad Kyeongheung Yun$^{2}$\quad Youngbin Choi$^{1}$\\
  \textbf{Sehyun Hwang$^{2}$\quad Jungseul Ok$^{1,2}$\quad Dongwoo Kim$^{1,2}$} \\
  $^{1}$GSAI, POSTECH\hspace{0.2cm} $^{2}$CSE, POSTECH\hspace{0.2cm} \\
  {\small\texttt{\{jsheo12304,yuonsinsa,choi.youngbin,sehyun03,jungseul,dongwookim\}@postech.ac.kr}} \\
}

\begin{document}

\maketitle

\input{tex/0_abstract}
\input{tex/1_intro}
\input{tex/2_preliminary}
\input{tex/3_method}
\input{tex/4_experiments}

\input{tex/5_related_work}
\input{tex/6_conclusion}

%\newpage
\bibliographystyle{plainnat}
\bibliography{ref}

\newpage
\crefalias{section}{appendix}
\crefalias{subsection}{appendix}
\input{tex/999_appendix}

%\newpage
%\clearpage
%\input{tex/check_test.tex}

\end{document}

%% file: tex/macro.tex
\newcommand{\Dtr}{\mathcal{D}}                          % training dataset
\newcommand{\Dpool}{\mathcal{D}_{\mathrm{pool}}}        % candidate pool (selection)

\newcommand{\thhat}{\hat{\theta}}
\newcommand{\thout}[1]{\hat{\theta}_{\Dtr \setminus #1}}
\newcommand{\thin}[1]{\hat{\theta}_{\Dtr \cup #1}}
\newcommand{\thresp}[2]{\hat{\theta}(#1;\,#2)}        % response function; usage: \thresp{\epsilon}{S}

\newcommand{\loss}{\ell}                                % per-example loss (alias for cleanliness)
\newcommand{\Lloss}{L}                                  % empirical training loss
\newcommand{\target}{f}                                 % target function (alias)

\newcommand{\Dout}[1]{\mathcal{I}^{-}(#1)}            % LGO effect; usage: \Dout{S}
\newcommand{\Din}[1]{\mathcal{I}^{+}(#1)}                  % LGI effect; usage: \Din{S}
\newcommand{\Dhatout}[1]{\hat{\mathcal{I}}^{-}(#1)}   % LGO estimator; usage: \Dhatout{S}
\newcommand{\Dhatin}[1]{\hat{\mathcal{I}}^{+}(#1)}         % LGI estimator; usage: \Dhatin{S}
\newcommand{\DoutLin}[1]{\hat{\mathcal{I}}^{-}_{\mathrm{lin}}(#1)}
\newcommand{\DinLin}[1]{\hat{\mathcal{I}}^{+}_{\mathrm{lin}}(#1)}

\newcommand{\grad}{\nabla_\theta}                       % gradient w.r.t. theta
\newcommand{\Hess}{\nabla^2_\theta}                     % Hessian w.r.t. theta
\newcommand{\gex}[1]{g_{#1}}                            % per-example gradient; usage: \gex{i}
\newcommand{\Htr}{H}                                    % training-loss Hessian
\newcommand{\Hgn}{G}                                    % Gauss-Newton matrix
\newcommand{\Htgt}{H_{\target}}                         % target Hessian
\newcommand{\Mop}{M}                                    % curvature transfer operator H^{-1} H_f H^{-1}

\newcommand{\ushift}[1]{u_{#1}}                         % exact per-example shift; usage: \ushift{i}, \ushift{S}
\newcommand{\pair}[2]{\kappa(#1,\,#2)}                  % pairwise interaction; usage: \pair{z_i}{z_j}

\newcommand{\budget}{K}                                 % selection budget
\newcommand{\poolsize}{P}
\newcommand{\Ssel}[1]{S_{#1}}                           % selected subset after t iterations; usage: \Ssel{t}
\newcommand{\marg}[2]{m(#1 \mid #2)}                    % marginal score; usage: \marg{z_j}{S}

\newcommand{\Jac}[1]{J_{#1}}     % logit Jacobian
\newcommand{\KM}{K_{\Mop}}        % M-weighted NTK

\newcommand{\tabnum}[2]{{#1{\footnotesize±}{\scriptsize#2}}}
\newcommand{\besttabnum}[2]{{\textbf{#1}{\footnotesize\textbf{±}}{\scriptsize\textbf{#2}}}}

\DeclareMathOperator*{\argmin}{arg\,min}

\newtheorem*{remark}{Remark}

\newtheorem{proposition}{Proposition}
\crefname{appendix}{Appendix}{Appendices}
\Crefname{appendix}{Appendix}{Appendices}

%% file: tex/0_abstract.tex
\begin{abstract}
Influence functions approximate how removing a training example changes a quantity of interest, called the target function, such as a held-out loss. To estimate the influence of a group of examples, the standard practice is to sum the individual influences of its members. However, this sum does not capture how examples jointly affect the target: a pair of examples may be redundant or complementary, but the sum cannot distinguish these cases. We propose an interaction-aware influence function that characterizes how interactions between examples influence the target. By expanding the target to second order around the trained parameters, we obtain an estimator that augments the standard sum with a pairwise interaction term that captures the alignment between two examples' effects on the target. We empirically evaluate our estimator in two settings. First, on six dataset--model pairs spanning logistic regression, MLPs, and ResNet-9, our estimator tracks leave-group-out retraining substantially better than first-order influence across all settings. Second, when used as a greedy selection rule for instruction-tuning data on Llama-3.1-8B, it beats prior influence-based and representation-similarity baselines on five of seven downstream tasks, in a regime where standard influence-based selection underperforms random selection. 
%Code is available at \url{https://anonymous.4open.science/r/Interaction_IF-45D6}.
\end{abstract}

%% file: tex/1_intro.tex
\section{Introduction}
\label{sec:introduction}

% P1: Influence functions and the target function, with the two-stage framing
Quantifying how training examples shape a trained model is a foundational problem in machine learning, with applications including selecting valuable training data~\citep{engstrom2024dsdm,xia2024less,yu2024mates}, understanding what kinds of data benefit a model~\citep{grosse2023studying,heo2025influence}, and fairly compensating data providers~\citep{choe2024your,ghorbani2019data}. The conceptual gold standard is leave-one-out retraining: remove a training example, retrain the model, and measure how some quantity of interest has changed. Since retraining for every example is prohibitive, influence functions~\citep{koh2017understanding} estimate this counterfactual in closed form by modeling two things in turn: how reweighting a training example would shift the trained parameters, and how those shifted parameters would change the quantity of interest. The quantity of interest, called the \emph{target function}, is left to the user. It can be the loss on a held-out example~\citep{koh2017understanding}, an average over a validation set, the likelihood of a downstream task answer~\citep{choe2024your,xia2024less}, or any other smooth function of the model. This flexibility is what lets a single approximation serve so many applications.

% P2: Group attribution, additive limitation, and prior second-order work
Most of these applications, including subset selection, group-robustness analysis, and data valuation, ultimately ask about \emph{groups} of examples rather than single ones. The textbook extension scores a group as the sum of its members' individual influences~\citep{koh2019accuracy}, implicitly assuming examples contribute independently. This assumption is rarely true. Two near-duplicate examples each look highly influential on their own, yet adding both has roughly the same effect as adding one, so additive scoring double-counts the overlap and overestimates the group's true effect~\citep{hu2024most,huang2025approximations}. The same blind spot makes top-$k$ influence-based data selection produce subsets that are highly redundant rather than collectively useful~\citep{zhang2024harnessing}, since the most individually influential examples tend to be similar. \citet{basu2020second} previously addressed this with a second-order correction that makes the estimator non-additive, but their construction refines only how the trained parameters shift, missing how the target function responds to those shifts.

We argue that group attribution is fundamentally about how training examples affect the target. Therefore, interactions between examples should be characterized in terms of their joint effect on the target, not just on the parameter shift.
To this end, we propose an \emph{interaction-aware influence function} that augments the standard sum of individual influences with a pairwise interaction term over examples in the group. 
The pairwise term captures the joint effect on the target: it is positive when the two examples push the model in similar directions and negative when they push in opposite directions.
%The pairwise interaction is positive when the two examples' parameter shifts reinforce each other, and negative when they cancel.
% We further derive a closed-form factorization of the interaction term across classification settings, proving that two similar examples are jointly beneficial when their labels differ and redundant when their labels agree. This formally recovers classical principles in supervised learning, where similar examples with differing labels shape the decision boundary while those with matching labels provide overlapping evidence~\citep{birodkar2019semantic,cortes1995support,mirzasoleiman2020coresets,robinsoncontrastive}. 
We further derive a closed-form factorization of the interaction term across classification settings, proving that two similar examples are jointly complementary when their labels differ and redundant when their labels agree, formally recovering classical principles in supervised learning~\citep{birodkar2019semantic,cortes1995support,mirzasoleiman2020coresets,robinsoncontrastive}.
% As a natural application of the proposed influence function, we propose a data selection method that uses the interaction term to discount candidates whose effect overlaps with already-selected examples, addressing the redundancy of top-$k$ influence-based selection.
As an application, we use the interaction term to discount candidates whose effects overlap with already selected examples, addressing the redundancy in top-$k$ influence-based selection.

We evaluate the method in two settings. First, we test whether our estimator faithfully tracks leave-group-out retraining across six dataset--model pairs, by assessing Spearman rank correlation with ground-truth retraining effects. 
It substantially outperforms prior influence-based training data attribution methods, demonstrating the importance of accounting for interactions in group attribution. 
Second, we evaluate the data selection method at scales ranging from small MLPs to instruction-tuning of Llama-3.1-8B~\citep{grattafiori2024llama}. 
On small-scale models, our method selects subsets with class diversity comparable to random selection, whereas baselines collapse onto a few classes and fall below random. 
In instruction-tuning data selection, our method consistently improves over random selection and achieves the strongest overall performance on five of seven downstream tasks.

%% file: tex/2_preliminary.tex
\section{Preliminary}
\label{sec:preliminary}

We first review the classical influence function and its standard extension to group attribution. We summarize the notation used throughout the paper in \Cref{tab:notation} of \Cref{apx:notation}. 
Let \(\Dtr=\{z_i\}_{i=1}^N\) be the training set of size \(N\), where each example $z_i=(x_i,y_i)$ consists of an input $x_i$ and a label $y_i$, and let
\(\loss(z_i,\theta)\) denote the loss of example \(z_i\) at parameter \(\theta\in\Theta\).
For any subset \(A\subseteq \Dtr\), define the empirical risk and its minimizer as
\begin{equation}
\label{eq:empirical-risk}
    \mathcal R_A(\theta)
    \coloneqq
    \frac{1}{|A|}\sum_{z_i\in A}\loss(z_i,\theta),
    \qquad
    \hat\theta_A
    \coloneqq
    \arg\min_\theta \mathcal R_A(\theta).
\end{equation}
We write \(\thhat\coloneqq \hat\theta_{\Dtr}\) for the parameter trained on the full dataset. For a subset \(S\subseteq \Dtr\), the parameter obtained by retraining after removing \(S\) is \(\thout{S}\).

Let \(\target:\Theta\to\mathbb R\) be a differentiable scalar target function of the trained parameters that we treat as a quantity to be minimized, such as the loss on a test example or the average loss on a validation set.\footnote{A target to be maximized can be handled by negating $\target$.} The exact leave-group-out effect of removing \(S\) is
\begin{equation}
\label{eq:exact-lgo}
    \Dout{S}
    \coloneqq
    \target(\thout{S}) - \target(\thhat).
\end{equation}
For a singleton \(S=\{z_i\}\), this reduces to the usual leave-one-out effect.
Computing \(\Dout{S}\) exactly requires retraining the model without \(S\), which is expensive even for a single example and infeasible for many candidate groups.

Influence functions approximate this retraining effect through infinitesimal reweighting~\citep{koh2017understanding}. Instead of removing \(S\) directly, consider the parameter obtained after adding a small weight \(\epsilon\) to the losses of examples in \(S\):
\begin{equation}
\label{eq:response-function}
    \thresp{\epsilon}{S}
    \coloneqq
    \arg\min_\theta
    \left\{
        \frac{1}{N}\sum_{z_i\in\Dtr}\loss(z_i,\theta)
        +
        \epsilon\sum_{z_i\in S}\loss(z_i,\theta)
    \right\}.
\end{equation}
At \(\epsilon=0\), this recovers the original parameter \(\thhat\). Since each example has weight \(1/N\) in the empirical risk, setting \(\epsilon=-1/N\) removes the contribution of every example in \(S\). 
Thus, leave-group-out retraining can be viewed as a finite step along the reweighting path \(\thresp{\epsilon}{S}\).

The influence function replaces this finite step with two first-order approximations. Let $\Htr\coloneqq \tfrac{1}{N}\sum_{z_i\in\Dtr} \Hess \loss(z_i,\thhat)$ be the Hessian of the training objective. Assuming that $\loss(z_i,\cdot)$ is twice continuously differentiable for every $z_i\in\Dtr$ and that $\Htr$ is nonsingular at $\thhat$, the first approximation linearizes the reweighting path $\thresp{\epsilon}{S}$ in $\epsilon$ at $\epsilon=0$, yielding the parameter shift induced by removing $S$ as
\begin{equation}
\label{eq:parameter-shift}
    \thout{S} - \thhat
    \approx
    \frac{1}{N}\Htr^{-1}\sum_{z_i\in S}\grad \loss(z_i,\thhat).
\end{equation}
To simplify the notation, let $\ushift{i} \coloneqq \Htr^{-1} \grad \loss(z_i, \thhat)$ denote the per-example parameter shift induced by $z_i$, and let $\ushift{S} \coloneqq \sum_{z_i \in S} \ushift{i}$ denote its aggregate over $S$. 
The second approximation linearizes the target function $\target$ around $\thhat$, giving the standard group influence-function estimate of $\Dout{S}$:
\begin{equation}
\label{eq:standard-group-if}
    \DoutLin{S}
    \coloneqq
    \frac{1}{N}
    \grad \target(\thhat)^\top
    \ushift{S}.
    %\Htr^{-1}\sum_{z_i\in S}\grad \loss(z_i,\thhat).
\end{equation}
The derivations of \Cref{eq:parameter-shift} and \Cref{eq:standard-group-if} 
are provided in \Cref{apx:derivation_if}. Letting $\DoutLin{z_i}$ denote the 
singleton case, the standard group estimate is additive: 
$\DoutLin{S} = \sum_{z_i \in S} \DoutLin{z_i}$. This additivity is a 
consequence of the first-order approximation, not a property of the exact 
retraining effect $\Dout{S}$. As a result, standard influence functions 
cannot capture interactions among examples in a group, which motivates the 
interaction-aware approximation developed in the next section.

%% file: tex/3_method.tex
\section{Method}
\label{sec:method}
We now develop our interaction-aware influence function, which captures how examples in a group jointly affect a target function. \Cref{sec:method-attribution} derives a group-attribution estimator by expanding the target function to second order, yielding the standard additive influence plus a pairwise interaction term. \Cref{sec:method-selection} then extends the estimator to a greedy data-selection procedure that discounts candidates whose effect overlaps with already-selected examples. Full derivations and proofs for this section are provided in \Cref{apx:derivation_method}.

\input{tex/3_method/3_1_interaction_aware_influence_functions}
\input{tex/3_method/3_2_iterative_subset_selection}

%% file: tex/3_method/3_1_interaction_aware_influence_functions.tex
\subsection{Interaction-aware influence functions}
\label{sec:method-attribution}

The standard influence function loses interaction information because the first-order approximation treats each example's effect as independent of the others. We extend the first-order approximation to a second-order one to recover the interaction information.

\paragraph{Second-order expansion of the target function.}
We expand the target function $\target$ to second order around $\thhat$:
\begin{equation}
\label{eq:target-taylor}
\target(\thout{S}) - \target(\thhat)
\;=\;
\grad \target(\thhat)^\top \delta_S
\;+\;
\tfrac{1}{2}\, \delta_S^\top \Htgt\, \delta_S
\;+\; O(\|\delta_S\|^3),
\end{equation}
where $\delta_S \coloneqq \thout{S} - \thhat$ is the parameter shift induced by removing $S$ and $\Htgt \coloneqq \Hess \target(\thhat)$ is the Hessian of the target at $\thhat$. Substituting \Cref{eq:parameter-shift} into \Cref{eq:target-taylor} yields our interaction-aware influence function, an estimator of $\Dout{S}$:
\begin{equation}
\label{eq:second-order-influence}
\Dhatout{S}
\;\coloneqq\;
\underbrace{\frac{1}{N}\,\grad \target(\thhat)^\top \ushift{S}}_{\text{first-order influence}}
\;+\;
\underbrace{\frac{1}{2N^2}\, \ushift{S}^\top \Htgt\, \ushift{S}}_{\text{interaction term (ours)}}.
\end{equation}
The first term recovers the standard first-order estimate $\DoutLin{S}$ from \Cref{eq:standard-group-if}. The second term, which we call the interaction term, is a curvature correction induced by the target function and is the source of non-additivity across examples in $S$.

The same expansion applies to the data-addition setting, which arises in problems such as data selection and active learning. Adding $S$ to $\Dtr$ flips the sign of the linear term while leaving the quadratic term unchanged, giving
\begin{equation}
\label{eq:second-order-influence-in}
%\Din{S}
%\;\approx\;
\Dhatin{S}
\;\coloneqq\;
-\,\frac{1}{N}\,\grad \target(\thhat)^\top \ushift{S}
\;+\;
\frac{1}{2N^2}\, \ushift{S}^\top \Htgt\, \ushift{S}.
\end{equation}
The full derivation for the data-addition setting is provided in \Cref{apx:derivation_addition}.
In practice, we use a damped Gauss--Newton Hessian approximation to ensure invertibility, and apply EK-FAC~\citep{george2018fast,grosse2023studying} and low-rank gradient projection~\citep{choe2024your} for scalable Hessian computation. Details of these practical approximations are provided in \Cref{apx:hessian_approximation}.

\paragraph{Interpreting the pairwise structure of the interaction term.}
The interaction term decomposes into a sum of pairwise contributions between examples in $S$:
\begin{equation}
\label{eq:interaction-pairwise}
\ushift{S}^\top \Htgt\, \ushift{S}
\;=\;
\sum_{z_i, z_j \in S} \pair{z_i}{z_j},
\qquad
\pair{a}{b} \coloneqq \ushift{a}^\top \Htgt\, \ushift{b}.
\end{equation}
For interpretive clarity, we first consider the case in which $\Htgt$ is positive definite; $\pair{a}{b}$ is then the inner product of the parameter shifts $\ushift{a}$ and $\ushift{b}$ under the metric induced by $\Htgt$. A positive value indicates that the two shifts are aligned under this metric, meaning the examples perturb the parameters in directions that produce similar changes in the target function, while orthogonal or opposing shifts produce values near zero or negative. In this sense, $\pair{a}{b}$ quantifies how \emph{similarly} two examples act on the target.
This similarity perspective explains why the additive approximation underestimates the exact retraining effect on groups of similar examples \citep{hu2024most}. Such groups produce large pairwise interactions that the additive approximation discards, leading to systematic underestimation. In the more general setting where $\Htgt$ is indefinite, $\pair{a}{b}$ generalizes from an inner product to a symmetric bilinear form. We provide a spectral analysis of this case in \Cref{apx:spectral_interpretation}.

\paragraph{Relation between the first-order and interaction terms.}
A concrete scenario clarifies how the two terms combine. Suppose we wish to identify a group $S$ whose removal would most reduce a target loss $\target$. The first-order term, which sums individual influences, picks out examples whose per-example influence $\tfrac{1}{N}\grad \target(\thhat)^\top \ushift{i}$ is negative, since each such removal individually lowers $\target$. On a group of near-duplicate examples of this kind, the additive sum predicts a large benefit, scaling linearly with group size. The interaction term, however, is non-negative whenever $\Htgt$ is positive semidefinite, and scales superlinearly with group size when the per-example shifts align. Adding it back therefore shrinks the predicted benefit, capturing the diminishing returns of removing redundant examples.
%and explaining the systematic underestimation of \citet{hu2024most}.

In the addition setting, the linear term flips sign while the quadratic term is unchanged. In this case, the two terms genuinely trade off: the linear term rewards candidates whose addition would individually lower $\target$, capturing quality, while the interaction term penalizes similarity within the chosen group, capturing redundancy. 
This trade-off resonates with a recurring principle in data selection~\citep{zhang2024harnessing} and active learning~\citep{ash2019deep,citovsky2021batch,kirsch2019batchbald}: effective subsets pair individually helpful examples with non-overlapping ones, typically by combining a per-example score with an explicit diversity term. Here, the same balance emerges directly from a second-order expansion of the target, rather than from an externally imposed design.
The greedy procedure in \Cref{sec:method-selection} exploits this trade-off, balancing per-example quality against group diversity at every step.

\paragraph{Interpreting the pairwise interaction in binary logistic regression.}
To better understand the pairwise interaction, we analyze a tractable case in which the interaction admits a closed form: binary logistic regression with $\ell_2$ regularization. Let $\sigma$ denote the sigmoid function, and write $\sigma_i \coloneqq \sigma(\thhat^\top x_i)$ for the predicted probability on example~$i$. 
\begin{proposition}[Closed-form factorization of $\kappa$]
\label{prop:factorization}
For binary logistic regression at $\thhat$ with $y_i\in \{0,1\}$,
\begin{equation*}
\pair{a}{b} = (\sigma_a - y_a)(\sigma_b - y_b) \cdot \langle x_a, x_b \rangle_{\Mop},
\end{equation*}
where $\langle u, v \rangle_{\Mop} \coloneqq u^\top \Mop\, v$ is the bilinear form induced by $\Mop \coloneqq \Htr^{-1} \Htgt\, \Htr^{-1}$.
\end{proposition}
The proof relies on the closed-form gradient $\grad \loss(z_i, \thhat) = (\sigma_i - y_i) x_i$. Full details are given in \Cref{apx:proof_factorization}. 

\Cref{prop:factorization} shows that when $\langle x_a, x_b \rangle_{\Mop} > 0$, so that $x_a$ and $x_b$ are similar under the bilinear form induced by $\Mop$, the sign of $\pair{a}{b}$ is determined by the sign of the class-agreement factor $(\sigma_a - y_a)(\sigma_b - y_b)$. Since $\sigma_i \in [0, 1]$, this factor is positive for same-class pairs and negative for cross-class pairs. Because the target is a quantity to be minimized, this implies that adding two similar examples with different labels benefits the classifier beyond their individual effects, while adding similar examples with the same label yields diminishing returns.

This formally recovers two well-known principles of supervised learning: cross-class pairs with similar features shape the decision boundary and are thus complementary~\citep{cortes1995support,robinsoncontrastive,shrivastava2016training}, whereas same-class pairs with similar features provide overlapping evidence and yield diminishing returns~\citep{birodkar2019semantic,mirzasoleiman2020coresets, tonevaempirical}. \Cref{prop:factorization} shows that the interaction term encodes both principles directly, with the sign of $\pair{a}{b}$ aligning with the role each pair plays in shaping the classifier.
For clarity of exposition, we have presented this analysis in the binary logistic regression setting. Both the factorization and these conclusions hold for deep multi-class classifiers, as we develop in \Cref{apx:deep_extension}.

\begin{remark}[Comparison with \citet{basu2020second}]
Our framework expands the target function to second order, while the same second-order treatment can equally be applied to the reweighted minimizer $\hat\theta(\epsilon; S)$. \citet{basu2020second} pursue this alternative for group attribution. The two approaches yield structurally distinct estimators: ours decomposes into pairwise contributions between examples in $S$, which enables the similarity interpretation and the closed-form analysis in \Cref{prop:factorization}, while theirs operates at the parameter level without an analogous pairwise structure. The two approaches also differ in stability: $\hat\theta(\epsilon; S)$ need not be a smooth function of $\epsilon$ in non-convex settings, so its second-order expansion in $\epsilon$ is poorly conditioned, while the target functions we consider, such as the held-out loss, are twice continuously differentiable in $\theta$ and avoid this source of ill-conditioning. Consistent with this analysis, \citet{basu2020second} themselves report degraded performance of their estimator on non-linear models, a finding that aligns with our empirical comparison in \Cref{sec:exp_attribution}.
\end{remark}

%% file: tex/3_method/3_2_iterative_subset_selection.tex
\subsection{Interaction-aware data selection}
\label{sec:method-selection}

\input{fig/src/alg_greedy_selection}

We now propose a data selection method that exploits the interaction term to select examples from a candidate pool for addition to the training set. This problem reduces directly to group attribution: finding the group that most reduces the target when added to the training data. Formally, given a model trained on $\Dtr$ and a candidate pool $\Dpool$, we seek a subset $S \subseteq \Dpool$ of size $\budget$ that minimizes the following objective:
\begin{equation}
S^\star \;\coloneqq\; \argmin_{|S| = \budget} \target\!\left(\thin{S}\right),
\label{eq:selection-objective}
\end{equation}
where $\thin{S}$ denotes the minimizer of the empirical risk on the augmented 
set $\Dtr \cup S$. Since $\target(\thhat)$ does not depend on $S$, subtracting 
it from the objective preserves the minimizer, so this is equivalent to 
minimizing the inclusion effect $\Din{S} \coloneqq \target(\thin{S}) - 
\target(\thhat)$. As $\Din{S}$ is intractable to evaluate exactly, we use its 
estimator $\Dhatin{S}$ from \Cref{eq:second-order-influence-in} as our 
selection criterion.

\paragraph{Greedy procedure.}
Exact combinatorial optimization of $\Dhatin{S}$ is intractable, so we instead adopt a greedy procedure that, starting from $S = \varnothing$, repeatedly appends the candidate $z_i$ minimizing $\Dhatin{S\cup \{z_i\}}$. We quantify this change through the marginal score of adding $z_i \notin S$ to a current subset $S$:
\begin{equation}
\label{eq:marginal-definition}
\marg{z_i}{S} \;\coloneqq\; \Dhatin{S \cup \{z_i\}} - \Dhatin{S}.
\end{equation}
Substituting \Cref{eq:second-order-influence-in} into \Cref{eq:marginal-definition} and using $\ushift{S \cup \{z_i\}} = \ushift{S} + \ushift{i}$, which follows from the definition of $\ushift{S}$, to expand the quadratic term gives
\begin{equation}
\marg{z_i}{S}
\;\approx\;
-\frac{1}{N}\,\grad \target(\thhat)^\top \ushift{i}
\;+\;
\frac{1}{N^2}\,\ushift{S}^\top \Htgt\, \ushift{i}
\;+\;
\frac{1}{2N^2}\,\ushift{i}^\top \Htgt\, \ushift{i}.
\label{eq:marginal}
\end{equation}
The first term is the standard first-order influence of $z_i$, which captures its individual contribution to reducing the target. 
The second term is the key novelty of our procedure: it accumulates the pairwise interactions analyzed in \Cref{sec:method-attribution} between the candidate $z_i$ and each example in the already-selected subset $S$. When $z_i$ is similar to examples in $S$, the second term takes a large positive value and increases the marginal score. The procedure thereby avoids selecting such candidates. 
The third term depends only on the candidate $z_i$ itself, measuring how strongly its own parameter shift would perturb the target. 
At each iteration we append the candidate minimizing $\marg{z_i}{S}$ until $|S| = \budget$; the full procedure is given in \Cref{alg:greedy}.

The interaction-aware structure of \Cref{eq:marginal} sets our procedure apart from selection based on the first-order influence~\citep{xia2024less}, which chooses the top-$k$ examples by individual influence score. Since high-influence examples tend to share characteristics~\citep{zhang2024harnessing}, this top-$k$ rule yields redundant subsets and consequently diminishing returns. Our second term addresses this limitation by penalizing candidates similar to already-selected examples, thereby promoting diversity.

\paragraph{Complexity.}
The per-candidate quantities $\{\ushift{i}\}$, $\{w_i \coloneqq \Htgt\, \ushift{i}\}$, and $\{q_i \coloneqq \ushift{i}^\top w_i\}$ in \Cref{alg:greedy} depend on the candidate pool and target $\target$ but not on the iterates $S$, so we precompute them once before the greedy loop. Each iteration then reduces to $O(\poolsize)$ inner products in the dimension $d$ of $\ushift{i}$ plus a single update of the accumulator $w$, yielding a total cost of $O(\budget \poolsize d)$. 
Wall-clock measurements are reported in \Cref{apx:compute}.

%% file: fig/src/alg_greedy_selection.tex
\begin{algorithm}[t]
\caption{Greedy selection with interaction-aware influence}
\label{alg:greedy}
\begin{algorithmic}[1]
\REQUIRE Candidate pool $\Dpool = \{z_i\}_{i=1}^{\poolsize}$, budget $\budget$, precomputed $\{\ushift{i}\}$, $\{w_i \coloneqq \Htgt\, \ushift{i}\}$, $\{q_i \coloneqq \ushift{i}^\top w_i\}$, target gradient $\grad \target(\thhat)$, training set size $N \coloneqq |\Dtr|$
\ENSURE Selected subset $S$ with $|S| \le \budget$
\STATE Initialize $S \leftarrow \emptyset$ and accumulator $w \leftarrow \mathbf{0}$, where $w$ tracks $\Htgt\, \ushift{S}$
\FOR{$t = 1, \ldots, \budget$}
    \FOR{$i \notin S$}
        \STATE $\marg{z_i}{S} \leftarrow -\frac{1}{N}\grad \target(\thhat)^\top \ushift{i} + \frac{1}{N^2}\,w^\top \ushift{i} + \frac{1}{2N^2}\,q_i$
    \ENDFOR
    \STATE $i^\star \leftarrow \arg\min_{i \notin S} \marg{z_i}{S}$
    \STATE $S \leftarrow S \cup \{z_{i^\star}\}$, \quad $w \leftarrow w + w_{i^\star}$
\ENDFOR
\RETURN $S$
\end{algorithmic}
\end{algorithm}

%% file: tex/4_experiments.tex
\input{tex/4_experiments/4_1_estimating_group_influence}
\input{tex/4_experiments/4_2_training_data_subset_selection}

%% file: tex/4_experiments/4_1_estimating_group_influence.tex
\section{Faithfulness to Retraining}
\label{sec:exp_attribution}
We validate the interaction-aware estimator developed in \Cref{sec:method-attribution} along two axes. We first measure how faithfully each estimator tracks the leave-group-out retraining effect on small-scale models, where exact ground truth can be computed, and compare against first-order influence and prior attribution methods. 
We then examine the pairwise interaction term's behavior across class pairs 
to illustrate the class-agreement structure of \Cref{prop:factorization}.

\paragraph{Setup.}
We report the Spearman rank correlation between estimated and ground-truth group influences. The ground-truth $\Dout{S}$ is obtained by retraining from scratch on $\Dtr \setminus S$ and measuring the change in average held-out test loss. This evaluation is analogous to the linear datamodeling score (LDS)~\citep{park2023trak}, but we avoid the term to prevent ambiguity, as our attribution is not a linear combination of individual scores.

\input{fig/src/fig_exp_group_attribution}

We consider six dataset--model pairs spanning classification (MNIST~\citep{lecun2002gradient}, FashionMNIST~\citep{xiao2017fashion}, CIFAR-10~\citep{krizhevsky2009learning}) and regression (Concrete~\citep{yeh1998modeling}, Parkinsons~\citep{tsanas2009accurate}), with models ranging from logistic regression (LR) to MLPs and ResNet-9~\citep{he2016deep}. To evaluate our estimator under conditions that are challenging for group attribution, we construct groups with high intra-group similarity, where existing group estimators are known to break down~\citep{hu2024most}. 

We compare our practical estimator against three influence-based methods. Throughout, we use \emph{F} to denote the first-order influence term, \emph{B} for the second-order term from \citet{basu2020second}, and \emph{I} for our interaction term. The baselines are then the first-order influence function~\citep{koh2017understanding} (\emph{F}), its second-order variant~\citep{basu2020second} (\emph{F+B}), and our estimator augmented with the Basu term (\emph{F+I+B}). Our method corresponds to \emph{F+I}. We additionally compare against two attribution methods aggregated to the group level by summing individual scores: TRAK~\citep{park2023trak} and TracIn~\citep{pruthi2020estimating}. 
Group construction, training, and influence computation details are provided 
in \Cref{apx:experiments}.

\input{fig/src/fig_exp_class-wise_helpful_harmful}%
\paragraph{Results.}
\Cref{fig:group_attribution} reports the Spearman rank correlation across the six dataset--model pairs.
The accuracy of first-order influence (\emph{F}) varies substantially across settings, ranging from strong correlation on MNIST to near-zero on the harder pairs.
Our estimator (\emph{F+I}) achieves the highest correlation in every setting, improving upon first-order influence by up to $0.67$ and yielding meaningfully positive correlations even where first-order influence performs poorly. This supports our claim that the interaction term captures information that first-order influence systematically overlooks.
The second-order variant (\emph{F+B}) shows instability, empirically confirming the ill-conditioning of expanding $\hat\theta(\epsilon; S)$ discussed in \Cref{sec:method-attribution}. 
The same correction applied to our estimator (\emph{F+I+B}) yields either a negligible change or a substantial drop.
TRAK and TracIn perform comparably to first-order influence. 
These findings are robust to the damping hyperparameter, as shown in \Cref{apx:additional_exp}.

\paragraph{Pairwise interaction analysis across classes.}
Given a multiclass classification problem, we identify the class pairs whose joint effect on the test loss differs most from the sum of their individual influences. For each pair of classes $(i, j)$, including within-class pairs $i = j$, we compute the average pairwise interaction $\kappa$ over all example pairs with one example in class $i$ and the other in class $j$. \Cref{fig:class_pairs} shows representative images from the class pairs with the lowest and highest averages for MNIST/LR, FashionMNIST/MLP, and CIFAR-10/ResNet-9. The lowest-$\kappa$ pairs are visually similar examples from different classes, while the highest-$\kappa$ pairs come from the same class, consistent with \Cref{prop:factorization}.

% \Cref{prop:factorization} shows that visually similar examples are jointly beneficial when they belong to different classes and jointly redundant when they belong to the same class, reflecting their respective roles in shaping the decision boundary. 
% % \Cref{prop:factorization} shows that our estimator predicts that visually similar examples are beneficial when they belong to different classes and yield diminishing returns when they belong to the same class.
% We empirically verify this pattern across architectures as follows. For each pair of classes $(c, c')$, we sample examples from each class and compute the class-pair average interaction $\bar\kappa(c, c')$ as the mean of $\kappa(a, b)$ over sampled pairs with $y_a = c$ and $y_b = c'$.
% We then identify the most beneficial and most redundant class pairs, corresponding to the most negative and most positive $\bar\kappa$ respectively, on MNIST/LR, FashionMNIST/MLP, and CIFAR-10/ResNet-9.
% % We then identify the class pairs that our estimator predicts as the most benefical and the most redundant.
% \Cref{fig:class_pairs} confirms this pattern in every setting: across all three model-dataset pairs, the most beneficial class pairs are visually similar yet distinct, while the most redundant pairs are within-class.

%% file: fig/src/fig_exp_group_attribution.tex
\begin{figure}[t]
    \centering
    \includegraphics[width=\linewidth]{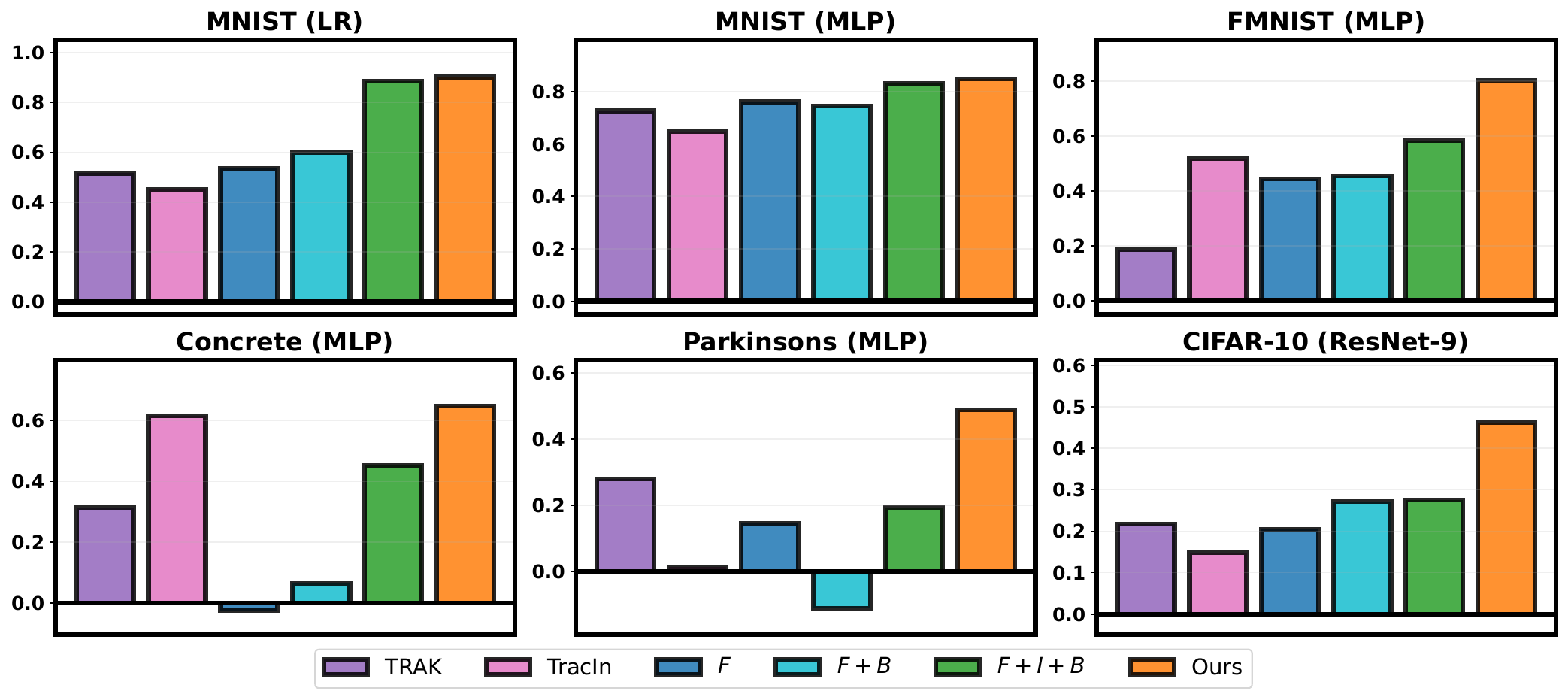}
    \caption{Spearman rank correlation between estimated and ground-truth group influences across six dataset--model pairs, where \emph{F}, \emph{B}, and \emph{I} denote the first-order influence function, the second-order term from \citet{basu2020second}, and our interaction term, respectively.
    %Red values show our improvement over first-order influence.
    }
    \label{fig:group_attribution}
\end{figure}

%% file: fig/src/fig_exp_class-wise_helpful_harmful.tex
\begin{wrapfigure}{r}{0.38\linewidth}
    \centering
    \vspace{-\baselineskip}
    \includegraphics[width=\linewidth]{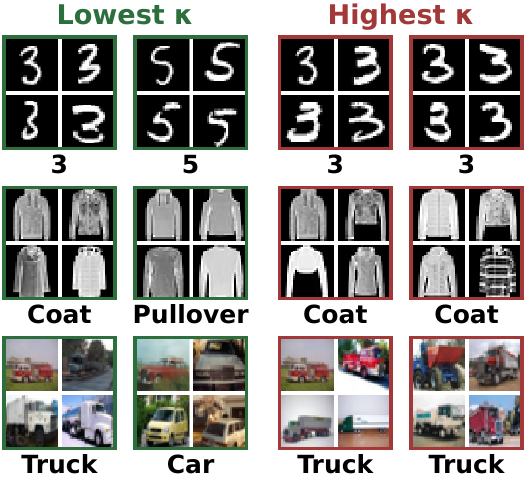}
    \caption{Representative images from the class pairs with the lowest (left) and highest (right) average pairwise interaction.}
    \label{fig:class_pairs}
    \vspace{-2\baselineskip}
\end{wrapfigure}

%% file: tex/4_experiments/4_2_training_data_subset_selection.tex
\section{Application to Subset Selection}
\label{sec:exp_selection}
We now evaluate our estimator as a selection criterion via \Cref{alg:greedy}. \Cref{sec:exp_validation_small_scale} validates the procedure at a scale where ground-truth subset quality can be measured by retraining. \Cref{sec:exp_llm_selection} then tests whether the gains scale to instruction-tuning data selection for Llama-3.1-8B~\citep{grattafiori2024llama}.

\subsection{Validation on Small-Scale Models}
\label{sec:exp_validation_small_scale}
This subsection isolates the contribution of the interaction term to selection quality. We compare against influence-function variants on the held-out test loss after retraining, and additionally analyze the diversity of the selected subsets through their class composition.

\paragraph{Setup.}
We apply \Cref{alg:greedy} to two-layer MLPs on MNIST and FashionMNIST. For each dataset, we construct a candidate pool $\Dpool$ of $|\Dpool|=5{,}000$ examples sampled uniformly at random from the training set.
For each selection method, we form a subset $S\subseteq\Dpool$ of size $K\in\{500,1000,\ldots,5000\}$, retrain the MLP from scratch on $S$, and report the resulting held-out test loss. To diagnose selection diversity, we additionally report the class entropy $\mathcal{H}(S) \coloneqq -\sum_c p_c(S) \log p_c(S)$ of the selected subset, where $p_c(S)$ is the fraction of examples in $S$ belonging to class $c$, with higher values indicating more uniform class coverage. We compare our method (\emph{F+I}) against the same influence-function baselines as in \Cref{sec:exp_attribution} (\emph{F}, \emph{F+B}, \emph{F+I+B}) and random selection. Further details are in \Cref{apx:experiments}.

\paragraph{Results.}
\input{fig/src/fig_exp_small_data_selection}
The left and middle plots in \Cref{fig:small_selection} report the held-out test loss attained by retraining on each method's selected subset. Our method (\emph{F+I}) outperforms every baseline on both datasets and at every selection size, indicating that the interaction term translates into substantially better subsets. The first-order influence function (\emph{F}) degrades sharply and falls below random selection. We attribute this to the redundancy of the subsets it selects and analyze it in detail in the class-entropy analysis below. The second-order variant of \citet{basu2020second} (\emph{F+B}) improves over \emph{F} but still trails random selection at most budgets.

The right plot of \Cref{fig:small_selection} reports the class entropy of subsets selected by each method. At small budgets, both \emph{F} and \emph{F+B} collapse onto a few classes, whereas our method matches the entropy of random selection throughout, confirming that the interaction term penalizes redundant candidates. A subset that omits entire classes cannot train a competitive classifier, which directly accounts for the test-loss degradation observed above. The corresponding plot for FashionMNIST is provided in \Cref{apx:additional_exp}.

\subsection{Data Selection for LLM Instruction Tuning}
\label{sec:exp_llm_selection}
We now evaluate our selection method at LLM scale by fine-tuning Llama-3.1-8B on selected instruction-tuning subsets and measuring downstream task performance.

\paragraph{Setup.}
We use the LESS pool~\citep{xia2024less} of approximately 270K instruction-tuning examples as the candidate pool. Following standard practice in influence-based data selection for LLMs~\citep{xia2024less}, we adopt a warmup-then-select pipeline: we first fine-tune Llama-3.1-8B on a small random subset to obtain a warmup checkpoint, which serves as $\thhat$ in our framework. 
We then compute per-example projected gradients $\ushift{j}$ and target-curvature factors $\Htgt\, \ushift{j}$ at this checkpoint using LoGra~\citep{choe2024your}, and run \Cref{alg:greedy} to select $\budget = 13{,}534$ examples, amounting to $5\%$ of the pool, independently for each target task.
The target function $\target$ is the instruction-masked likelihood, evaluated on the official validation set when available and on a held-out split of the training set otherwise. We then re-finetune from the original pretrained checkpoint on the selected subset. Training details, including warmup configuration, optimizer, LoRA setup, damping, and per-method runtimes, are provided in \Cref{apx:experiments}.

We evaluate on seven downstream tasks covering three reasoning capabilities. For mathematical reasoning, we use GSM8K~\citep{cobbe2021training} and AQuA~\citep{ling2017program}. For commonsense and science reasoning, we use ARC-Easy~\citep{clark2018think}, HellaSwag~\citep{zellers2019hellaswag}, PIQA~\citep{bisk2020piqa}, and ECQA~\citep{aggarwal2021explanations}. For reading comprehension, we use SQuAD~\citep{rajpurkar2016squad}. We report exact match for GSM8K, F1 for SQuAD, and accuracy for the multiple-choice tasks. 

We compare against two influence-based baselines, two representation-similarity baselines, and a random baseline. The influence-based baselines select examples with the highest target-influence scores: Additive IF uses the standard first-order influence function~\citep{koh2017understanding}, sharing our pipeline and differing only in the absence of the interaction term. LESS~\citep{xia2024less} uses cosine similarity between low-rank projections of training and target gradients. The representation-similarity baselines select examples whose feature embeddings most closely match those of the target task: RDS+~\citep{ivisondata2025} uses pretrained-LM hidden states as features, and NV-Embed~\citep{lee2024nv} uses embeddings from a retrieval-trained LLM. Random selection samples $\budget$ examples uniformly from the pool. 

\input{fig/src/tab_llm_selection}
\paragraph{Results.}
\Cref{tab:llm_selection} reports the mean and standard deviation over five seeds. Our method attains the best performance on five of the seven tasks and surpasses random selection on every task. The two influence-based baselines, Additive IF and LESS, are outperformed on six of the seven tasks; Additive IF, which ablates our interaction term, even falls below random on most tasks, isolating the contribution of interaction-aware selection. LESS and NV-Embed further drop to below $9\%$ accuracy on AQuA, whereas our method remains stable across all tasks. 
On GSM8K, although our method outperforms random selection and the influence-based baselines, the representation-similarity baselines perform especially well. We conjecture this stems from GSM8K's narrow design: problems are restricted to elementary arithmetic, single integer answers, and short, simple descriptions. This concentrated distribution plays to the strength of representation-similarity selection, which directly retrieves semantically near-identical examples from the pool.

%% file: fig/src/fig_exp_small_data_selection.tex
\begin{figure}[t]
    \centering
    \includegraphics[width=\linewidth]{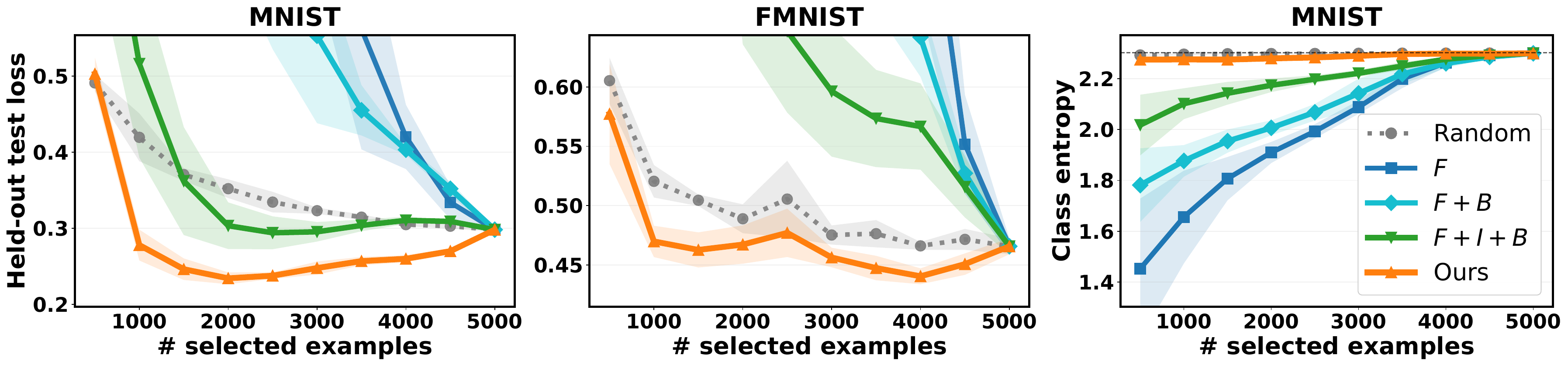}
    \caption{Held-out test loss after retraining on the selected subset on MNIST (left) and FashionMNIST (middle); class entropy of the subset on MNIST (right). Across selection sizes, lines and shaded regions show the mean and standard deviation over five seeds.}
    \label{fig:small_selection}
\end{figure}

%% file: fig/src/tab_llm_selection.tex
\begin{table*}[t]
\centering
\small
\caption{Test performance of fine-tuned Llama-3.1-8B on seven target tasks. We report the mean and standard deviation over five seeds; the best result in each column is shown in \textbf{bold}.}
\label{tab:llm_selection}
\resizebox{\textwidth}{!}{
\begin{tabular}{l ccccccc c}
\toprule
\textbf{Method} & GSM8K & AQuA & ARC-E & HellaSwag & PIQA & ECQA & SQuAD & \textbf{Avg.} \\
\midrule
Random            & \tabnum{46.75}{2.69} & \tabnum{24.65}{0.72} & \tabnum{90.03}{0.19} & \tabnum{56.46}{0.69} & \tabnum{77.50}{0.47} & \tabnum{69.85}{0.22} & \tabnum{75.76}{0.55} & 63.00 \\
Additive IF       & \tabnum{19.35}{1.07} & \tabnum{20.63}{1.03} & \tabnum{88.72}{0.30} & \tabnum{48.66}{1.99} & \tabnum{77.38}{0.79} & \tabnum{69.52}{0.29} & \tabnum{76.92}{0.30} & 57.31 \\
LESS              & \tabnum{41.83}{1.32} & \tabnum{ 8.82}{2.87} & \tabnum{90.11}{0.15} & \tabnum{62.11}{0.87} & \tabnum{77.40}{0.49} & \tabnum{71.19}{0.12} & \tabnum{76.73}{0.51} & 61.17 \\
RDS+              & \besttabnum{58.44}{1.29} & \tabnum{24.65}{2.87} & \tabnum{88.70}{0.32} & \tabnum{52.19}{1.03} & \tabnum{75.14}{1.14} & \tabnum{70.88}{0.41} & \tabnum{57.60}{1.91} & 61.09 \\
NV-Embed          & \tabnum{57.97}{0.47} & \tabnum{ 8.43}{1.06} & \tabnum{89.01}{0.16} & \tabnum{53.94}{0.74} & \tabnum{75.13}{0.44} & \besttabnum{71.75}{0.24} & \tabnum{76.52}{1.10} & 61.82 \\
\midrule
\textbf{Ours} & \tabnum{52.69}{0.96} & \besttabnum{30.94}{0.72} & \besttabnum{90.44}{0.11} & \besttabnum{63.43}{0.34} & \besttabnum{79.39}{0.32} & \tabnum{71.17}{0.19} & \besttabnum{79.87}{0.34} & \textbf{66.85} \\
\bottomrule
\end{tabular}
}
\end{table*}

%% file: tex/5_related_work.tex
\section{Related Work}
\paragraph{Data attribution and influence functions.}
Data attribution methods quantify the contribution of individual training examples to a model's learned parameters or predictions on a target~\citep{ghorbani2019data,koh2017understanding,park2023trak,pruthi2020estimating}. Among these, influence functions~\citep{koh2017understanding} approximate this counterfactual via local linearization, with subsequent work improving scalability through random-projection approaches~\citep{park2023trak}, trajectory-based methods~\citep{pruthi2020estimating}, efficient curvature approximations~\citep{grosse2023studying,schioppa2022scaling,wang2025better}, and extensions to large language and generative models~\citep{agarwal2026neural,chang2025scalable,choe2024your,kwon2023datainf,mlodozeniec2025influence}. Complementary work studies the fragility of influence estimates~\citep{basu2020influence,deng2024versatileinfluencefunctiondata,kreer2026bayesian,rubinstein2026rescaled}. \citet{ye2025towards} also expand the validation loss to second order, but only for single-example influence in noisy-label settings, without group attribution. Prior work also shows that summing individual attributions mismeasures group-level effects~\citep{hu2024most,koh2019accuracy,saunshi2022understanding}. Closest to our setting, \citet{basu2020second} address this additive gap via a response-side expansion; we instead expand the target function, yielding closed-form pairwise interactions. \citet{wang2024data} and \citet{wei2025ji2s} also model joint influence but focus on trajectory-accumulated rather than counterfactual effects.

\paragraph{Training data selection.}
Data selection methods score examples by influence~\citep{agarwal2026neural,choe2024your,dai2025improving,xia2024less}, diversity~\citep{pan2024g}, learned utility~\citep{engstrom2024dsdm,yu2024mates}, or Shapley-style contributions~\citep{ghorbani2019data}. Scalable variants leverage influence distillation or small-model trajectories~\citep{nikdan2026efficient,yang2024smalltolarge}. Redundancy-aware pruning further penalizes similar examples via pairwise similarity objectives~\citep{tan2025data}. However, individually influential examples need not form the best subset once interactions make collective effects non-additive~\citep{hu2024most,huang2025approximations,wang2024rethinking,yu2026grouplevel}. Our second-order expansion of the target function enables interaction-aware selection by updating each candidate's marginal utility against the already-selected set.

%% file: tex/6_conclusion.tex
\section{Conclusion}
\label{sec:conclusion}
We presented an interaction-aware influence function obtained from a second-order expansion of the target function, decomposing group attribution into a standard first-order term and a pairwise interaction term. Empirically, our method improves Spearman correlation with ground-truth retraining effects on six dataset--model pairs. In subset selection, it outperforms random selection on every task and beats existing influence-based and representation-similarity baselines on five of seven downstream tasks. Two limitations suggest natural directions for future work: our greedy selection performs strongly in practice but lacks formal optimality guarantees, and our estimator inherits approximation error from the underlying Hessian approximations. Advances in scalable inverse-Hessian estimation would directly improve robustness.

%% file: tex/999_appendix.tex
\appendix\
\input{tex/999_appendix/A_notation}
\input{tex/999_appendix/B_preliminary_derivations}
\input{tex/999_appendix/C_method_derivation}
\input{tex/999_appendix/D_Hessian_approximations}

\input{tex/999_appendix/E_experimental_settings}
\input{tex/999_appendix/F_additional_experiments}

%% file: tex/999_appendix/A_notation.tex
\section{Notation}
\label{apx:notation}

\Cref{tab:notation} consolidates the notation used throughout the paper. The symbols are grouped into five categories: data and model objects, group-level effects on the target function, parameter shifts and curvature matrices, the pairwise interaction term, and quantities specific to data selection. Within each category, definitions are listed in the order they first appear in the main text, and we use the same symbols in the appendix derivations.

\begin{table}[h]
\centering
\caption{Summary of notation used throughout the main text and appendices, organized by category.}
\label{tab:notation}
\small
\begin{tabular}{@{}p{0.32\linewidth}p{0.6\linewidth}@{}}
\toprule
\textbf{Symbol} & \textbf{Description} \\
\midrule
\multicolumn{2}{@{}l@{}}{\textbf{Data and model}} \\
\addlinespace[2pt]
$\Dtr = \{z_i\}_{i=1}^{N}$ & Training dataset of $N$ examples \\
$z_i = (x_i, y_i)$ & $i$-th training example with input $x_i$ and label $y_i$ \\
$S \subseteq \Dtr$ & Subset (group) of training examples \\
$\theta \in \mathbb{R}^p$ & Model parameters \\
$\thhat$ & Parameters obtained by training on $\Dtr$ \\
$\thout{S}$ & Parameters obtained by retraining on $\Dtr \setminus S$ \\
$\thin{S}$ & Parameters obtained by retraining on $\Dtr \cup S$ \\
$\thresp{\epsilon}{S}$ & Parameters when examples in $S$ are upweighted by $\epsilon$ \\
$\loss(z, \theta)$ & Per-example loss \\
$\Lloss(\theta) = \tfrac{1}{N}\sum_{i} \loss(z_i, \theta)$ & Empirical training loss \\
$\target(\theta)$ & Target function (e.g., loss on a held-out example) \\
\addlinespace
\midrule
\multicolumn{2}{@{}l@{}}{\textbf{Group-level effects}} \\
\addlinespace[2pt]
$\Dout{S}$ & Removal effect: $\target(\thout{S}) - \target(\thhat)$ \\
$\Din{S}$ & Inclusion effect: $\target(\thin{S}) - \target(\thhat)$ \\
$\DoutLin{S}$ & First-order estimate of $\Dout{S}$ \\
$\DinLin{S}$ & First-order estimate of $\Din{S}$ \\
$\Dhatout{S}$ & Our interaction-aware estimator of $\Dout{S}$ \\
$\Dhatin{S}$ & Our interaction-aware estimator of $\Din{S}$ \\
\addlinespace
\midrule
\multicolumn{2}{@{}l@{}}{\textbf{Parameter shifts and curvature}} \\
\addlinespace[2pt]
$\gex{i} \coloneqq \grad \loss(z_i, \thhat)$ & Per-example gradient at $\thhat$ \\
$\Htr \coloneqq \Hess \Lloss(\thhat)$ & Training-loss Hessian at $\thhat$ \\
$\Hgn$ & Gauss--Newton Hessian of the training loss at $\thhat$ \\
%$\Hhat \coloneqq \Hgn + \lambda I$ & Damped Gauss--Newton approximation of $\Htr$ \\
$\Htgt \coloneqq \Hess \target(\thhat)$ & Target-function Hessian at $\thhat$ \\
$\Mop \coloneqq \Htr^{-1} \Htgt \Htr^{-1}$ & Bilinear form arising in \Cref{prop:factorization} \\
$\ushift{i} \coloneqq \Htr^{-1} \gex{i}$ & Per-example parameter shift \\
$\ushift{S} \coloneqq \sum_{z_i \in S} \ushift{i}$ & Aggregate parameter shift over $S$ \\
%$\uhat{i} \coloneqq \Hhat^{-1} \gex{i}$ & Gauss--Newton-approximated per-example shift \\
%$\uhat{S} \coloneqq \sum_{z_i \in S} \uhat{i}$ & Aggregate approximated shift over $S$ \\
\addlinespace
\midrule
\multicolumn{2}{@{}l@{}}{\textbf{Pairwise interaction term}} \\
\addlinespace[2pt]
$\pair{z_i}{z_j} \coloneqq \ushift{i}^\top \Htgt \ushift{j}$ & Pairwise interaction between $z_i$ and $z_j$ \\
\addlinespace
\midrule
\multicolumn{2}{@{}l@{}}{\textbf{Data selection}} \\
\addlinespace[2pt]
$\Dpool$ & Candidate pool of examples \\
$\budget$ & Selection budget (target subset size) \\
$\Ssel{t}$ & Selected subset after $t$ greedy iterations \\
$\marg{z_j}{S}$ & Marginal score of adding $z_j$ to current subset $S$ \\
\bottomrule
\end{tabular}
\end{table}

%% file: tex/999_appendix/B_preliminary_derivations.tex
\section{Derivation of the First-Order Influence Function}
\label{apx:derivation_if}
We derive \Cref{eq:parameter-shift} and \Cref{eq:standard-group-if}, the first-order Taylor approximations of (i) the parameter shift induced by removing $S$ and (ii) the resulting change in the target function $f$.
Throughout, we maintain the assumptions stated in \Cref{sec:preliminary}: $\ell(z_i,\cdot)$ is twice continuously differentiable for every $z_i\in\mathcal D$, and the training-loss Hessian
\[
H \coloneqq \frac{1}{N}\sum_{z_i\in\mathcal D}\nabla_\theta^2\ell(z_i,\hat\theta)
\]
is nonsingular at $\hat\theta$, and the target function $f$ is differentiable at $\hat\theta$.
 
\paragraph{Proof strategy.}
Our goal is to approximate the target change $f(\hat\theta_{\mathcal D\setminus S}) - f(\hat\theta)$ caused by removing $S$, which we obtain in two first-order Taylor-expansion steps.
The first step approximates the parameter shift $\hat\theta_{\mathcal D\setminus S} - \hat\theta$: as noted in \Cref{sec:preliminary}, this parameter shift corresponds to a finite perturbation from $\epsilon=0$ to $\epsilon=-1/N$ along the reweighting path $\hat\theta(\epsilon;S)$, and we approximate it by a first-order Taylor expansion of $\hat\theta(\epsilon;S)$ in $\epsilon$, yielding \Cref{eq:parameter-shift}.
The second step substitutes the resulting parameter shift into a first-order Taylor expansion of $f$ around $\hat\theta$, yielding \Cref{eq:standard-group-if}.
The only quantity we need to compute for the first step is the slope $d\hat\theta(\epsilon;S)/d\epsilon$ at $\epsilon=0$.
To compute it, we use the fact that $\hat\theta(\epsilon;S)$ minimizes the perturbed objective for every $\epsilon$, so the gradient of the perturbed objective at $\hat\theta(\epsilon;S)$ remains zero as $\epsilon$ varies.
Differentiating this zero-gradient identity in $\epsilon$ yields a closed form for the slope.
 
\paragraph{Setup.}
Recall the perturbed objective from \Cref{eq:response-function},
\begin{equation}
    \mathcal R(\theta;\epsilon,S)
    \coloneqq
    \frac{1}{N}\sum_{z_i\in\mathcal D}\ell(z_i,\theta)
    +
    \epsilon\sum_{z_i\in S}\ell(z_i,\theta),
\label{eq:perturbed-objective}
\end{equation}
and its minimizer $\hat\theta(\epsilon;S) \coloneqq \arg\min_\theta \mathcal R(\theta;\epsilon,S)$.
At $\epsilon=0$ the perturbation vanishes and we recover $\hat\theta(0;S)=\hat\theta$.
At $\epsilon=-1/N$, substituting into \Cref{eq:perturbed-objective} cancels the loss terms for $z_i\in S$:
\[
    \mathcal R\bigl(\theta;\,-\tfrac{1}{N},\,S\bigr)
    = \frac{1}{N}\sum_{z_i\in\mathcal D}\ell(z_i,\theta) - \frac{1}{N}\sum_{z_i\in S}\ell(z_i,\theta)
    = \frac{1}{N}\sum_{z_i\in\mathcal D\setminus S}\ell(z_i,\theta).
\]
This differs from the empirical risk $\mathcal R_{\mathcal D\setminus S}(\theta) = \frac{1}{|\mathcal D\setminus S|}\sum_{z_i\in\mathcal D\setminus S}\ell(z_i,\theta)$ defined in \Cref{sec:preliminary} only by a positive constant factor, so its minimizer coincides with the leave-group-out parameter $\hat\theta_{\mathcal D\setminus S}$.
 
\paragraph{Zero-gradient identity at the minimizer.}
Since $\hat\theta(\epsilon;S)$ is a minimizer of $\mathcal R(\cdot;\epsilon,S)$, the gradient of $\mathcal R$ in $\theta$ is zero at $\hat\theta(\epsilon;S)$.
Concretely, computing $\nabla_\theta \mathcal R(\theta;\epsilon,S)$ from \Cref{eq:perturbed-objective} and evaluating at $\theta = \hat\theta(\epsilon;S)$ gives
\begin{equation}
    \frac{1}{N}\sum_{z_i\in\mathcal D}\nabla_\theta\ell\bigl(z_i,\hat\theta(\epsilon;S)\bigr)
    +
    \epsilon\sum_{z_i\in S}\nabla_\theta\ell\bigl(z_i,\hat\theta(\epsilon;S)\bigr)
    = 0.
\label{eq:optimality}
\end{equation}
This identity holds for every $\epsilon$ in a neighborhood of $0$. By the implicit function theorem, the nonsingularity of $H$ further ensures that the minimizer path $\hat\theta(\epsilon;S)$ is differentiable in $\epsilon$ near $\epsilon=0$, so we may differentiate both sides of the identity in $\epsilon$ to extract the desired slope $d\hat\theta(\epsilon;S)/d\epsilon|_{\epsilon=0}$.
 
\paragraph{Differentiating in $\epsilon$.}
Each per-example gradient $\nabla_\theta\ell(z_i, \hat\theta(\epsilon;S))$ depends on $\epsilon$ only through $\hat\theta(\epsilon;S)$.
Differentiating \Cref{eq:optimality} in $\epsilon$ and applying the chain rule to each such term gives
\begin{equation}
    \underbrace{\left[
        \frac{1}{N}\sum_{z_i\in\mathcal D}\nabla_\theta^2\ell\bigl(z_i,\hat\theta(\epsilon;S)\bigr)
        + \epsilon\sum_{z_i\in S}\nabla_\theta^2\ell\bigl(z_i,\hat\theta(\epsilon;S)\bigr)
    \right]}_{\text{Hessian of }\mathcal R(\cdot;\epsilon,S)}
    \frac{d\hat\theta(\epsilon;S)}{d\epsilon}
    +
    \sum_{z_i\in S}\nabla_\theta\ell\bigl(z_i,\hat\theta(\epsilon;S)\bigr)
    = 0.
\label{eq:differentiated-optimality}
\end{equation}
At $\epsilon=0$, the second summand inside the bracket is zero due to its $\epsilon$ prefactor, and the bracket reduces to $H$.
Using $g_i \coloneqq \nabla_\theta\ell(z_i,\hat\theta)$ as defined in \Cref{sec:preliminary}, \Cref{eq:differentiated-optimality} becomes
\begin{equation}
    H\cdot\frac{d\hat\theta(\epsilon;S)}{d\epsilon}\bigg|_{\epsilon=0}
    +
    \sum_{z_i\in S}g_i = 0.
\end{equation}
Since $H$ is invertible, we solve for the slope:
\begin{equation}
    \frac{d\hat\theta(\epsilon;S)}{d\epsilon}\bigg|_{\epsilon=0}
    = -H^{-1}\sum_{z_i\in S}g_i.
\label{eq:sensitivity}
\end{equation}
 
\paragraph{Parameter-shift approximation (\Cref{eq:parameter-shift}).}
We now use the slope in \Cref{eq:sensitivity} inside the first Taylor expansion promised in the strategy.
Expanding $\hat\theta(\epsilon;S)$ to first order around $\epsilon=0$ gives
\begin{equation}
    \hat\theta(\epsilon;S)
    \approx
    \hat\theta + \epsilon\cdot\frac{d\hat\theta(\epsilon;S)}{d\epsilon}\bigg|_{\epsilon=0}
    =
    \hat\theta - \epsilon\,H^{-1}\sum_{z_i\in S}g_i.
\end{equation}
Setting $\epsilon=-1/N$, which corresponds to leaving $S$ out, recovers \Cref{eq:parameter-shift}:
\begin{equation}
    \hat\theta_{\mathcal D\setminus S} - \hat\theta
    \approx
    \frac{1}{N}H^{-1}\sum_{z_i\in S}g_i.
\label{eq:lgo-shift-derived}
\end{equation}
 
\paragraph{Target-change approximation via the parameter shift (\Cref{eq:standard-group-if}).}
For the second step, a first-order Taylor expansion of the target $f$ around $\hat\theta$ gives
\begin{equation}
    f(\hat\theta_{\mathcal D\setminus S}) - f(\hat\theta)
    \approx
    \nabla_\theta f(\hat\theta)^\top
    \bigl(\hat\theta_{\mathcal D\setminus S} - \hat\theta\bigr).
\end{equation}
Substituting the parameter shift from \Cref{eq:lgo-shift-derived} yields the standard first-order group influence estimate of \Cref{eq:standard-group-if}:
\begin{equation}
    \DoutLin{S}
    \coloneqq
    \frac{1}{N}\nabla_\theta f(\hat\theta)^\top H^{-1}\sum_{z_i\in S}g_i.
\end{equation}
The additivity follows immediately, since the right-hand side depends on $S$ only through the linear sum $\sum_{z_i\in S}g_i$.

%% file: tex/999_appendix/C_method_derivation.tex
\section{Derivations and Proofs for the Interaction-Aware Influence Function}
\label{apx:derivation_method}

\input{tex/999_appendix/C_method_derivation/C_1_interaction_aware_IF}

\input{tex/999_appendix/C_method_derivation/C_2_data_addition}
\input{tex/999_appendix/C_method_derivation/C_3_interaction_decomposition}
\input{tex/999_appendix/C_method_derivation/C_4_spectral_decomposition}
\input{tex/999_appendix/C_method_derivation/C_5_proposition_1}
\input{tex/999_appendix/C_method_derivation/C_6_deep_extension}
\input{tex/999_appendix/C_method_derivation/C_7_marginal_score}

%% file: tex/999_appendix/C_method_derivation/C_1_interaction_aware_IF.tex
\subsection{Derivation of the interaction-aware influence function}
\label{apx:derivation_second_order}

We derive the interaction-aware influence function in \Cref{eq:second-order-influence} by combining the second-order Taylor expansion of the target function with the first-order parameter shift induced by removing $S$.

Recall from \Cref{eq:parameter-shift} in \Cref{sec:preliminary} that the parameter shift induced by removing $S$ admits the first-order approximation
\begin{equation}
\label{eq:apx-parameter-shift}
\delta_S \;\coloneq\; \thout{S} - \thhat
\;\approx\;
\frac{1}{N}\,\Htr^{-1} \sum_{z_i \in S} \gex{i}
\;=\;
\frac{1}{N}\, \ushift{S},
\end{equation}
where the last equality uses $\ushift{i} \coloneq \Htr^{-1} \gex{i}$ and $\ushift{S} \coloneq \sum_{z_i \in S} \ushift{i}$. The second-order Taylor expansion of $\target$ around $\thhat$ in \Cref{eq:target-taylor} is
\begin{equation}
\label{eq:apx-target-taylor}
\target(\thout{S}) - \target(\thhat)
\;=\;
\grad \target(\thhat)^\top \delta_S
\;+\;
\tfrac{1}{2}\, \delta_S^\top \Htgt\, \delta_S
\;+\; O(\|\delta_S\|^3).
\end{equation}

Substituting \Cref{eq:apx-parameter-shift} into the linear term of \Cref{eq:apx-target-taylor} gives
\begin{equation}
\label{eq:apx-linear-term}
\grad \target(\thhat)^\top \delta_S
\;\approx\;
\frac{1}{N}\, \grad \target(\thhat)^\top \ushift{S}.
\end{equation}
Substituting the same approximation into the quadratic term gives
\begin{equation}
\label{eq:apx-quadratic-term}
\tfrac{1}{2}\, \delta_S^\top \Htgt\, \delta_S
\;\approx\;
\tfrac{1}{2}\, \left(\frac{1}{N}\, \ushift{S}\right)^\top \Htgt\, \left(\frac{1}{N}\, \ushift{S}\right)
\;=\;
\frac{1}{2 N^2}\, \ushift{S}^\top \Htgt\, \ushift{S},
\end{equation}
where the factor $1/N^2$ arises from the two factors of $1/N$ in $\delta_S$. Combining \Cref{eq:apx-linear-term,eq:apx-quadratic-term} and recalling that the leave-group-out effect is $\Dout{S} = \target(\thout{S}) - \target(\thhat)$ yields
\begin{equation}
\label{eq:apx-second-order-influence}
\Dout{S}
\;\approx\;
\frac{1}{N}\, \grad \target(\thhat)^\top \ushift{S}
\;+\;
\frac{1}{2 N^2}\, \ushift{S}^\top \Htgt\, \ushift{S},
\end{equation}
which is \Cref{eq:second-order-influence} in the main text. The remainder term $O(\|\delta_S\|^3)$ is dropped, consistent with the second-order approximation. The first term recovers the standard first-order estimate $\DoutLin{S}$ from \Cref{eq:standard-group-if}, and the second term is the curvature-induced interaction term that gives rise to non-additivity across examples in $S$.

%% file: tex/999_appendix/C_method_derivation/C_2_data_addition.tex
\subsection{Derivation for the data-addition setting}
\label{apx:derivation_addition}

We now derive the addition-setting influence function in \Cref{eq:second-order-influence-in}. The argument parallels \Cref{apx:derivation_second_order}, with the only change appearing in the sign of the parameter shift induced by adding $S$ rather than removing it.

Consider the reweighted objective in \Cref{eq:response-function} with weight $\epsilon$ assigned to the examples in $S$. Setting $\epsilon = -1/N$ removes the contribution of $S$, recovering the leave-group-out parameter $\thout{S}$ as a finite step along the reweighting path; setting $\epsilon = +1/N$ instead adds an extra copy of each example in $S$ with weight $1/N$, which corresponds to retraining on the augmented set $\Dtr \cup S$ with each new example carrying the same per-example weight as in $\Dtr$. The first-order parameter shift induced by this addition is therefore
\begin{equation}
\label{eq:apx-parameter-shift-in}
\thin{S} - \thhat
\;\approx\;
-\frac{1}{N}\,\Htr^{-1} \sum_{z_i \in S} \gex{i}
\;=\;
-\frac{1}{N}\, \ushift{S},
\end{equation}
which differs from \Cref{eq:apx-parameter-shift} only in sign. A complete derivation through the implicit function theorem is given in \Cref{apx:derivation_if}; the sign flip reflects the opposite direction of the reweighting step.

Substituting \Cref{eq:apx-parameter-shift-in} into the second-order Taylor expansion of $\target$ around $\thhat$ yields
\begin{align}
\target(\thin{S}) - \target(\thhat)
&\;\approx\;
\grad \target(\thhat)^\top \!\left(-\frac{1}{N}\, \ushift{S}\right)
\;+\;
\tfrac{1}{2}\, \left(-\frac{1}{N}\, \ushift{S}\right)^\top \Htgt\, \left(-\frac{1}{N}\, \ushift{S}\right) \notag \\
&\;=\;
-\frac{1}{N}\, \grad \target(\thhat)^\top \ushift{S}
\;+\;
\frac{1}{2 N^2}\, \ushift{S}^\top \Htgt\, \ushift{S}.
\label{eq:apx-second-order-influence-in}
\end{align}
The linear term flips sign because the parameter shift itself is negated, while the quadratic term is invariant under this negation: the two factors of $-1/N$ in the bilinear form combine to a positive $1/N^2$. Recalling that $\Din{S} = \target(\thin{S}) - \target(\thhat)$ gives \Cref{eq:second-order-influence-in} in the main text.

This sign asymmetry has a direct consequence for our selection criterion in \Cref{sec:method-selection}. When $\target$ is to be minimized, a more negative $\Din{S}$ is preferable, so the linear term rewards candidates whose parameter shifts align with $-\grad \target(\thhat)$. The quadratic term, in contrast, contributes the same sign in both removal and addition settings, reflecting the fact that interactions among examples in $S$ depend on their joint geometry rather than on the direction of the perturbation.

%% file: tex/999_appendix/C_method_derivation/C_3_interaction_decomposition.tex
\subsection{Pairwise decomposition of the interaction term}
\label{apx:derivation_decompositions}

The interaction term $\ushift{S}^\top \Htgt\, \ushift{S}$ in \Cref{eq:second-order-influence,eq:second-order-influence-in} admits a pairwise decomposition used in \Cref{sec:method-attribution}, which we derive below. 

\paragraph{Pairwise decomposition.}
Recall that $\ushift{S} = \sum_{z_i \in S} \ushift{i}$ is the aggregate of per-example shifts. Substituting this definition into the bilinear form and expanding gives
\begin{equation}
\label{eq:apx-pairwise-decomposition}
\ushift{S}^\top \Htgt\, \ushift{S}
\;=\;
\left(\sum_{z_i \in S} \ushift{i}\right)^\top \Htgt \left(\sum_{z_j \in S} \ushift{j}\right)
\;=\;
\sum_{z_i \in S} \sum_{z_j \in S} \ushift{i}^\top \Htgt\, \ushift{j}
\;=\;
\sum_{z_i, z_j \in S} \pair{z_i}{z_j},
\end{equation}
where $\pair{a}{b} \coloneq \ushift{a}^\top \Htgt\, \ushift{b}$ as defined in \Cref{eq:interaction-pairwise}. This recovers the pairwise form used throughout \Cref{sec:method-attribution}.

\paragraph{Self and cross contributions.}
The double sum in \Cref{eq:apx-pairwise-decomposition} ranges over ordered pairs and can be further separated into diagonal and off-diagonal contributions:
\begin{equation}
\label{eq:apx-self-cross-decomposition}
\ushift{S}^\top \Htgt\, \ushift{S}
\;=\;
\underbrace{\sum_{z_i \in S} \pair{z_i}{z_i}}_{\text{self contribution}}
\;+\;
\underbrace{2 \sum_{\{z_i, z_j\} \subset S,\, i \neq j} \pair{z_i}{z_j}}_{\text{cross contribution}},
\end{equation}
where the cross contribution sums over unordered pairs $\{z_i, z_j\}$ with $i \neq j$ and the factor of two reflects the symmetry $\pair{z_i}{z_j} = \pair{z_j}{z_i}$ inherited from $\Htgt = \Htgt^\top$. The self contribution $\sum_i \pair{z_i}{z_i} = \sum_i \ushift{i}^\top \Htgt\, \ushift{i}$ is determined entirely by the individual examples and would persist even if examples in $S$ were processed independently. The cross contribution is the part of the interaction term that genuinely encodes interactions between distinct examples and is the source of non-additivity beyond the additive baseline. In the main text we present the unified double-sum form in \Cref{eq:interaction-pairwise} so that the analysis remains agnostic to whether self or cross terms dominate. The split into self and cross contributions in \Cref{eq:apx-self-cross-decomposition} becomes useful when isolating the genuinely interactional content, which we exploit in the marginal score derivation in \Cref{apx:derivation_marginal}.

%========================== Old =================================
\if\else
\subsection{Spectral and pairwise decompositions of the interaction term}
\label{apx:derivation_decompositions}

The interaction term $\ushift{S}^\top \Htgt\, \ushift{S}$ in \Cref{eq:second-order-influence,eq:second-order-influence-in} admits two equivalent decompositions used in \Cref{sec:method-attribution}. We derive each in turn.

\paragraph{Spectral decomposition.}
Since $\target$ is twice continuously differentiable at $\thhat$, the target Hessian $\Htgt = \Hess \target(\thhat)$ is symmetric and admits an eigendecomposition
\begin{equation}
\label{eq:apx-eigendecomposition-Hf}
\Htgt \;=\; \sum_k \mu_k\, v_k v_k^\top,
\end{equation}
where $\{\mu_k\}$ are the real eigenvalues and $\{v_k\}$ form an orthonormal basis of unit eigenvectors. Substituting \Cref{eq:apx-eigendecomposition-Hf} into the bilinear form $\ushift{S}^\top \Htgt\, \ushift{S}$ gives
\begin{equation}
\label{eq:apx-spectral-decomposition}
\ushift{S}^\top \Htgt\, \ushift{S}
\;=\;
\ushift{S}^\top \!\left(\sum_k \mu_k\, v_k v_k^\top\right)\! \ushift{S}
\;=\;
\sum_k \mu_k\, (v_k^\top \ushift{S})^\top (v_k^\top \ushift{S})
\;=\;
\sum_k \mu_k\, (v_k^\top \ushift{S})^2,
\end{equation}
which is \Cref{eq:eigendecomposition} in the main text. Each summand measures the squared projection of $\ushift{S}$ onto the eigendirection $v_k$, weighted by the corresponding curvature $\mu_k$. Eigendirections with $\mu_k > 0$ contribute positively to the interaction term and correspond to directions of local convexity of $\target$ at $\thhat$. Eigendirections with $\mu_k < 0$ contribute with the opposite sign.

\paragraph{Pairwise decomposition.}
Recall that $\ushift{S} = \sum_{z_i \in S} \ushift{i}$ is the aggregate of per-example shifts. Substituting this definition into the bilinear form and expanding gives
\begin{equation}
\label{eq:apx-pairwise-decomposition}
\ushift{S}^\top \Htgt\, \ushift{S}
\;=\;
\left(\sum_{z_i \in S} \ushift{i}\right)^\top \Htgt \left(\sum_{z_j \in S} \ushift{j}\right)
\;=\;
\sum_{z_i \in S} \sum_{z_j \in S} \ushift{i}^\top \Htgt\, \ushift{j}
\;=\;
\sum_{z_i, z_j \in S} \pair{z_i}{z_j},
\end{equation}
where $\pair{a}{b} \coloneq \ushift{a}^\top \Htgt\, \ushift{b}$ as defined in \Cref{eq:interaction-pairwise}. This recovers the pairwise form used throughout \Cref{sec:method-attribution}.

\paragraph{Self and cross contributions.}
The double sum in \Cref{eq:apx-pairwise-decomposition} ranges over ordered pairs and can be further separated into diagonal and off-diagonal contributions:
\begin{equation}
\label{eq:apx-self-cross-decomposition}
\ushift{S}^\top \Htgt\, \ushift{S}
\;=\;
\underbrace{\sum_{z_i \in S} \pair{z_i}{z_i}}_{\text{self contribution}}
\;+\;
\underbrace{2 \sum_{\{z_i, z_j\} \subset S,\, i \neq j} \pair{z_i}{z_j}}_{\text{cross contribution}},
\end{equation}
where the cross contribution sums over unordered pairs $\{z_i, z_j\}$ with $i \neq j$ and the factor of two reflects the symmetry $\pair{z_i}{z_j} = \pair{z_j}{z_i}$ inherited from $\Htgt = \Htgt^\top$. The self contribution $\sum_i \pair{z_i}{z_i} = \sum_i \ushift{i}^\top \Htgt\, \ushift{i}$ is determined entirely by the individual examples and would persist even if examples in $S$ were processed independently. The cross contribution is the part of the interaction term that genuinely encodes interactions between distinct examples and is the source of non-additivity beyond the additive baseline. In the main text we present the unified double-sum form in \Cref{eq:interaction-pairwise} so that the spectral and pairwise decompositions share the same starting expression and the analysis remains agnostic to whether self or cross terms dominate. The split into self and cross contributions in \Cref{eq:apx-self-cross-decomposition} becomes useful when isolating the genuinely interactional content, which we exploit in the marginal score derivation in \Cref{apx:derivation_marginal}.

The two decompositions in \Cref{eq:apx-spectral-decomposition,eq:apx-pairwise-decomposition} are equivalent reorganizations of the same quadratic form. The first groups contributions by eigendirection, while the second groups them by example pairs. Each perspective supports a different interpretation in the main text. \Cref{eq:apx-spectral-decomposition} clarifies the role of curvature in driving redundancy or reinforcement along principal directions, and \Cref{eq:apx-pairwise-decomposition} enables the closed-form analysis of $\pair{a}{b}$ given in \Cref{prop:factorization}.
\fi

%% file: tex/999_appendix/C_method_derivation/C_4_spectral_decomposition.tex
\subsection{Spectral interpretation of the pairwise interaction}
\label{apx:spectral_interpretation}
 
We provide a spectral interpretation of the pairwise interaction $\pair{a}{b}$ that holds in the general setting where the target Hessian $\Htgt$ may be indefinite, going beyond the inner-product reading of \Cref{sec:method-attribution}.
%The decomposition reveals two properties of $\pair{a}{b}$ that are not directly visible from the bilinear definition.
%First, each eigendirection of $\Htgt$ contributes \emph{independently} to $\pair{a}{b}$, with a sign and magnitude determined by both alignment and curvature.
%Second, this curvature-aware structure is what underlies the redundancy and complementarity behavior our estimator captures.
%We first read the spectral decomposition through a weighted-similarity perspective that recovers the inner-product picture of \Cref{sec:method-attribution}.
%We then turn to what changes when $\Htgt$ becomes indefinite.
 
\paragraph{Spectral decomposition of the interaction term.}
Since $\target$ is twice continuously differentiable at $\thhat$, $\Htgt$ is symmetric and admits the eigendecomposition
\begin{equation}
\label{eq:apx-eigendecomposition-Hf}
\Htgt \;=\; \sum_k \mu_k\, v_k v_k^\top,
\end{equation}
where $\{\mu_k\}$ are the real eigenvalues and $\{v_k\}$ form an orthonormal basis of unit eigenvectors.
Substituting \Cref{eq:apx-eigendecomposition-Hf} into the definition of the interaction term $\pair{a}{b} = \ushift{a}^\top \Htgt\, \ushift{b}$ yields
\begin{equation}
\label{eq:apx-spectral-decomposition}
\pair{a}{b}
\;=\;
\sum_k \mu_k\, (v_k^\top \ushift{a})(v_k^\top \ushift{b}).
\end{equation}
Each summand isolates the contribution of a single eigendirection $v_k$.
The product $(v_k^\top \ushift{a})(v_k^\top \ushift{b})$ measures how the parameter shifts of $a$ and $b$ are aligned along $v_k$.
The eigenvalue $\mu_k$ then specifies how strongly this alignment contributes to $\pair{a}{b}$, as well as its sign.
The spectral view thus provides additional information beyond the bilinear-form definition, allowing $\pair{a}{b}$ to be read as a sum of independent contributions, one per eigendirection of $\Htgt$.
 
\paragraph{Reading the decomposition as a weighted similarity score.}
A useful way to read \Cref{eq:apx-spectral-decomposition} is as an aggregate similarity score.
The shifts $\ushift{a}$ and $\ushift{b}$ are compared from multiple perspectives, one per eigendirection $v_k$.
These per-direction comparisons are then combined through a weighted sum, with the eigenvalues $\mu_k$ acting as importance weights.
%This is analogous to assessing how similar two people look by comparing them across several aspects, such as face, body shape, and profile, while prioritizing some aspects over others through different weights.
When $\Htgt$ is positive definite all weights $\mu_k$ are positive, so alignment along any eigendirection contributes consistently in the positive direction to the aggregate score.
This recovers the inner-product reading of \Cref{sec:method-attribution} in spectral form.
The additional insight is that the weight assigned to each perspective is precisely the curvature of $\target$ along that direction.
 
%\paragraph{Aggregate score and second-order target change.}
%Beyond its similarity reading, $\pair{a}{b}$ is itself the pairwise contribution to the second-order change of $\target$ induced by the joint perturbation of $a$ and $b$, as decomposed in \Cref{eq:apx-self-cross-decomposition} of \Cref{apx:derivation_decompositions}.
%It is a \emph{signed} correction to the additive first-order estimate.
%A positive value of $\pair{a}{b}$ increases the second-order target change relative to the sum of individual influences, while a negative value decreases it.
%Throughout this paper we treat $\target$ as a quantity to be minimized, so an increase in target change is detrimental.
%Under this convention, positive interactions are harmful and negative interactions are beneficial.
%In the positive-definite setting, every per-direction contribution shares the same sign as the alignment, so larger pairwise similarity translates into a larger positive correction.
%This corresponds to the harmful joint effect described in \Cref{sec:method-attribution}.
 
\paragraph{Extension to the indefinite case.}
When $\Htgt$ is instead indefinite, $\pair{a}{b}$ is no longer an inner product but a symmetric bilinear form.
The eigendecomposition in \Cref{eq:apx-eigendecomposition-Hf} remains valid, so the per-direction decomposition of $\pair{a}{b}$ in \Cref{eq:apx-spectral-decomposition} still applies.
The only change is that some eigenvalues $\mu_k$ may now be negative.
Concretely, suppose $\ushift{a}$ and $\ushift{b}$ are aligned along an eigendirection $v_k$, that is, $(v_k^\top \ushift{a})(v_k^\top \ushift{b}) > 0$:
\begin{itemize}
    \item if $\mu_k > 0$ (a locally convex direction), the per-direction contribution is positive, so this aligned perspective increases the second-order correction and is harmful under the loss-minimization convention;
    \item if $\mu_k < 0$ (a locally concave direction), the per-direction contribution is negative, so this aligned perspective decreases the second-order correction and is beneficial under the same convention.
\end{itemize}
The same logic applies with reversed signs when the two shifts are anti-aligned along $v_k$.
In that case, anti-alignment along a positive-curvature direction contributes negatively, while anti-alignment along a negative-curvature direction contributes positively.
The interaction term thus evaluates pairwise behavior in a curvature-aware manner.
The alignment between two examples can either drive $\target$ toward worse values or pull it toward better ones, depending on the sign of the curvature along that direction.

%% file: tex/999_appendix/C_method_derivation/C_5_proposition_1.tex
\subsection{Proof of \texorpdfstring{\Cref{prop:factorization}}{Proposition 1}}
\label{apx:proof_factorization}

We prove the closed-form factorization of $\pair{a}{b}$ stated in \Cref{prop:factorization}. We work in the binary logistic regression setting with $\ell_2$ regularization of strength $\beta > 0$, and let $\thhat$ denote the regularized empirical risk minimizer. The proof proceeds in three steps: deriving the closed-form per-example gradient and the corresponding Hessian of the regularized objective, expressing the per-example parameter shift in factored form, and substituting into the definition of $\pair{a}{b}$.

\paragraph{Step 1: per-example gradient and training-loss Hessian.}
For binary logistic regression, the per-example data loss takes the form
\begin{equation}
\label{eq:apx-logistic-loss}
\loss(z_i, \theta) \;=\; -\, y_i \log \sigma(\theta^\top x_i) \;-\; (1 - y_i) \log\!\left(1 - \sigma(\theta^\top x_i)\right),
\end{equation}
where $y_i \in \{0, 1\}$ is the binary label. Using the standard identity $\sigma'(t) = \sigma(t)(1 - \sigma(t))$ and writing $\sigma_i \coloneqq \sigma(\thhat^\top x_i)$, the per-example gradient evaluated at $\thhat$ is
\begin{equation}
\label{eq:apx-logistic-gradient}
\grad \loss(z_i, \thhat) \;=\; (\sigma_i - y_i)\, x_i.
\end{equation}

The training objective is the regularized empirical risk
\begin{equation}
\label{eq:apx-regularized-risk}
\mathcal{R}(\theta) \;=\; \frac{1}{N} \sum_i \loss(z_i, \theta) \;+\; \frac{\beta}{2} \|\theta\|^2.
\end{equation}
Differentiating once gives $\grad \mathcal{R}(\theta) = \frac{1}{N} \sum_i \grad \loss(z_i, \theta) + \beta\, \theta$. Differentiating once more yields the training-loss Hessian
\begin{equation}
\label{eq:apx-training-hessian}
\Htr \;\coloneqq\; \Hess \mathcal{R}(\thhat)
\;=\; \frac{1}{N} \sum_i \Hess \loss(z_i, \thhat) \;+\; \beta\, I
\;=\; \frac{1}{N} \sum_i \sigma_i (1 - \sigma_i)\, x_i x_i^\top \;+\; \beta\, I,
\end{equation}
where the last equality uses the closed-form per-example Hessian of the binary logistic loss. The data-dependent part $\frac{1}{N} \sum_i \sigma_i(1 - \sigma_i)\, x_i x_i^\top$ is positive semidefinite as a sum of rank-one outer products with non-negative weights, and the regularizer contributes $\beta I$ with $\beta > 0$. Their sum $\Htr$ is therefore positive definite and hence invertible. We use \Cref{eq:apx-logistic-gradient} as the per-example gradient driving the influence approximation throughout the remainder of the proof.

\paragraph{Step 2: per-example parameter shift.}
Substituting the closed-form gradient into the definition $\ushift{i} \coloneqq \Htr^{-1} \grad \loss(z_i, \thhat)$ gives
\begin{equation}
\label{eq:apx-logistic-ushift}
\ushift{i} \;=\; \Htr^{-1} \!\left[(\sigma_i - y_i) x_i\right] \;=\; (\sigma_i - y_i)\, \Htr^{-1} x_i,
\end{equation}
where the second equality uses the linearity of matrix-vector multiplication and the fact that $(\sigma_i - y_i)$ is a scalar. Each per-example parameter shift therefore factors into a scalar prediction residual $(\sigma_i - y_i)$ and a vector $\Htr^{-1} x_i$ that depends only on the input feature $x_i$ through the inverse training-loss Hessian.

\paragraph{Step 3: factorization of $\pair{a}{b}$.}
Substituting \Cref{eq:apx-logistic-ushift} into the definition $\pair{a}{b} \coloneqq \ushift{a}^\top \Htgt\, \ushift{b}$ from \Cref{eq:interaction-pairwise} gives
\begin{align}
\pair{a}{b}
&\;=\; \left[(\sigma_a - y_a)\, \Htr^{-1} x_a\right]^\top \Htgt \left[(\sigma_b - y_b)\, \Htr^{-1} x_b\right] \notag \\
&\;=\; (\sigma_a - y_a)(\sigma_b - y_b)\, x_a^\top \Htr^{-1} \Htgt\, \Htr^{-1} x_b \notag \\
&\;=\; (\sigma_a - y_a)(\sigma_b - y_b)\, \langle x_a, x_b \rangle_{\Mop},
\label{eq:apx-kappa-factorization}
\end{align}
where the second equality factors the two scalar residuals out of the bilinear form and uses the symmetry of $\Htr^{-1}$, which holds because $\Htr$ is symmetric and positive definite, and the third equality applies the definition $\Mop \coloneqq \Htr^{-1} \Htgt\, \Htr^{-1}$ together with $\langle u, v \rangle_{\Mop} \coloneqq u^\top \Mop\, v$. This is exactly the factorization claimed in \Cref{prop:factorization}, completing the proof.

%% file: tex/999_appendix/C_method_derivation/C_6_deep_extension.tex
\subsection{Extension of \texorpdfstring{\Cref{prop:factorization}}{Proposition 1} to deep classifiers}
\label{apx:deep_extension}

\Cref{prop:factorization} establishes a closed-form factorization of $\pair{a}{b}$ for binary logistic regression. We now extend this result to deep classifiers, treating binary and multi-class classification in turn. The argument is in both cases a direct consequence of chain rule applied to the per-example gradient: the same scalar-residual--input-feature factorization that drives the proof of \Cref{prop:factorization} survives when the input feature $x_i$ is replaced by the logit Jacobian
\begin{equation}
\label{eq:apx-deep-jacobian}
\Jac{i} \;\coloneqq\; \grad f_{\thhat}(x_i),
\end{equation}
where $f_\theta : \mathcal{X} \to \mathbb{R}^C$ is the network logit output, with $C = 1$ for binary classification and $C \geq 2$ for multi-class classification. The bilinear-form matrix $\Mop \coloneqq \Htr^{-1} \Htgt\, \Htr^{-1}$ from \Cref{prop:factorization} is unchanged. We treat the binary case first as it offers the cleanest correspondence with \Cref{prop:factorization}, then state and prove the multi-class generalization.

\paragraph{Binary case.}
For binary classification ($C = 1$), $\Jac{i} \in \mathbb{R}^p$ is a vector. Writing $\sigma_i \coloneqq \sigma(f_{\thhat}(x_i))$, the per-example loss
$\loss(z_i, \theta) = -\, y_i \log \sigma(f_\theta(x_i)) - (1 - y_i) \log\!\left(1 - \sigma(f_\theta(x_i))\right)$
has gradient at $\thhat$ given by chain rule as
\begin{equation}
\label{eq:apx-deep-binary-gradient}
\grad \loss(z_i, \thhat) \;=\; (\sigma_i - y_i)\, \Jac{i},
\end{equation}
which differs from \Cref{eq:apx-logistic-gradient} only in that the input feature $x_i$ is replaced by the logit Jacobian $\Jac{i}$. Steps 2 and 3 of the proof of \Cref{prop:factorization} therefore carry over verbatim with $x_i \mapsto \Jac{i}$, yielding
\begin{equation}
\label{eq:apx-deep-binary-factorization}
\pair{a}{b} \;=\; (\sigma_a - y_a)(\sigma_b - y_b)\, \langle \Jac{a}, \Jac{b} \rangle_{\Mop}.
\end{equation}
The class-agreement scalar is unchanged because it depends only on the network's predictions and the labels. The input-similarity term $\langle x_a, x_b\rangle_\Mop$ generalizes to a Jacobian bilinear form $\langle \Jac{a}, \Jac{b}\rangle_\Mop$, which is an $\Mop$-weighted empirical neural tangent kernel evaluated at $\thhat$ when $\Htgt$ is positive semidefinite.

\paragraph{Decomposing the Jacobian bilinear form via the last layer.}
Writing $\theta = (\psi, w)$ for the feature-extractor parameters and the last-layer weight, and decomposing $f_\theta(x) = w^\top \phi_\psi(x)$ where $\phi_\psi : \mathcal{X} \to \mathbb{R}^{d_\phi}$ denotes the learned feature representation, the logit Jacobian $\Jac{i}$ admits the block decomposition
\begin{equation}
\label{eq:apx-jacobian-decomposition}
\Jac{i} = \begin{bmatrix} J_\phi(x_i)^\top w \\ \phi(x_i) \end{bmatrix}, \qquad J_\phi(x_i) \coloneqq \frac{\partial \phi_\psi(x_i)}{\partial \psi},
\end{equation}
corresponding to gradients with respect to $\psi$ (top block) and $w$ (bottom block). Substituting into the Jacobian bilinear form exposes both the learned feature representation and the feature Jacobian directly:
\begin{equation}
\label{eq:apx-binary-decomposed}
\langle \Jac{a}, \Jac{b}\rangle_\Mop \;=\; \begin{bmatrix} J_\phi(x_a)^\top w \\ \phi(x_a) \end{bmatrix}^\top \Mop \begin{bmatrix} J_\phi(x_b)^\top w \\ \phi(x_b) \end{bmatrix}.
\end{equation}
In \Cref{eq:apx-binary-decomposed}, the bottom-block component $\phi(x_i)$ takes the place of the raw input feature $x_i$ in \Cref{prop:factorization}. Since the network's representation is well known to map visually similar examples to similar $\phi(x_i)$, such examples contribute to $\pair{a}{b}$ through this block in the same way that input similarity contributes in the LR setting. The top-block component $J_\phi(x_i)^\top w$ captures how the feature representation responds to feature-extractor parameter perturbations along the readout direction $w$, an additional structure unique to deep classifiers that has no counterpart in the LR setting.

\paragraph{Multi-class case.}
For multi-class classification with $C \geq 2$, $\Jac{i} \in \mathbb{R}^{p \times C}$ is a matrix whose $c$-th column is the gradient of the $c$-th logit with respect to all parameters. Write $p_i \in \Delta^{C-1}$ for the softmax of $f_{\thhat}(x_i)$ and $y_i \in \{0,1\}^C$ for the one-hot label. The cross-entropy loss
$\loss(z_i, \theta) = -\sum_{c=1}^C y_{i,c} \log p_{i,c}(\theta)$
has gradient at $\thhat$ given by chain rule as
\begin{equation}
\label{eq:apx-deep-multiclass-gradient}
\grad \loss(z_i, \thhat) \;=\; \Jac{i}\, r_i,
\qquad
r_i \;\coloneqq\; p_i - y_i \;\in\; \mathbb{R}^C.
\end{equation}
\Cref{eq:apx-deep-multiclass-gradient} generalizes \Cref{eq:apx-logistic-gradient} in two parallel ways: the scalar residual $(\sigma_i - y_i)$ becomes the vector residual $r_i$, and the input feature $x_i$ becomes the matrix Jacobian $\Jac{i}$.

Substituting \Cref{eq:apx-deep-multiclass-gradient} into the definition $\ushift{i} \coloneqq \Htr^{-1}\, \grad \loss(z_i, \thhat)$ gives $\ushift{i} = \Htr^{-1} \Jac{i}\, r_i$. Plugging this into the pairwise interaction $\pair{a}{b} \coloneqq \ushift{a}^\top \Htgt\, \ushift{b}$ from \Cref{eq:interaction-pairwise} and factoring yields
\begin{align}
\pair{a}{b}
&\;=\; (\Htr^{-1} \Jac{a}\, r_a)^\top \Htgt\, (\Htr^{-1} \Jac{b}\, r_b) \notag \\
&\;=\; r_a^\top\, \Jac{a}^\top \Htr^{-1} \Htgt\, \Htr^{-1}\, \Jac{b}\, r_b \notag \\
&\;=\; r_a^\top\, \KM(x_a, x_b)\, r_b,
\label{eq:apx-deep-multiclass-factorization}
\end{align}
where the second equality uses the symmetry of $\Htr^{-1}$, and the third equality defines the matrix-valued bilinear form
\begin{equation}
\label{eq:apx-multiclass-ntk}
\KM(x_a, x_b) \;\coloneqq\; \Jac{a}^\top \Mop\, \Jac{b} \;\in\; \mathbb{R}^{C \times C},
\end{equation}
which is a matrix-valued empirical neural tangent kernel when $\Htgt$ is positive semidefinite. The factorization in \Cref{eq:apx-deep-multiclass-factorization} is the multi-class generalization of \Cref{eq:apx-kappa-factorization}: the scalar residual product $(\sigma_a - y_a)(\sigma_b - y_b)$ is replaced by the bilinear form $r_a^\top \KM\, r_b$, and the scalar feature similarity $\langle x_a, x_b\rangle_\Mop$ is replaced by the matrix-valued bilinear form $\KM(x_a, x_b)$. The binary case in \Cref{eq:apx-deep-binary-factorization} is recovered when $C = 1$, with $r_i = \sigma_i - y_i$ and $\KM = \langle \Jac{a}, \Jac{b}\rangle_\Mop$ a scalar.

\paragraph{Connection to the class-agreement structure.}
\Cref{prop:factorization} identifies cross-class similar pairs as beneficial and same-class similar pairs as redundant, with the sign of $\pair{a}{b}$ controlled by the residual product $(\sigma_a - y_a)(\sigma_b - y_b)$. The multi-class factorization in \Cref{eq:apx-deep-multiclass-factorization} preserves the structural form \emph{residual structure} $\times$ \emph{Jacobian similarity}, with the residual product replaced by the bilinear form $r_a^\top \KM\, r_b$. A direct calculation yields the residual-alignment identity
\begin{equation}
\label{eq:apx-residual-alignment}
r_a^\top r_b \;=\; \langle p_a, p_b\rangle - p_{a, y_b} - p_{b, y_a} + \mathds{1}[y_a = y_b],
\end{equation}
in which the indicator term contributes $+1$ for same-class pairs and $0$ for cross-class pairs. For predictions biased toward their true classes, this indicator drives the sign of the residual alignment positive for same-class pairs and negative for cross-class pairs; for example, with $p_{i,c} = (1-\delta)\,\mathds{1}[c = y_i] + \delta/(C-1)$ on the remaining classes, a short calculation gives $r_a^\top r_b = \delta^2\, C/(C-1) > 0$ for same-class pairs and $r_a^\top r_b = -\delta^2\, C/(C-1)^2 < 0$ for cross-class pairs. When two examples are similar so that $\Jac{a} \approx \Jac{b}$ and $\Mop$ is positive semidefinite, $\KM(x_a, x_b) \approx \Jac{a}^\top \Mop \Jac{a}$, which is symmetric PSD, and $\pair{a}{b}$ inherits the sign of the residual alignment. This recovers the cross-class-beneficial and same-class-redundant structure of \Cref{prop:factorization} and is consistent with the \Cref{fig:class_pairs} observation that this class-pair structure carries over from logistic regression to MLPs and ResNet-9.

%% file: tex/999_appendix/C_method_derivation/C_7_marginal_score.tex
\subsection{Derivation of the marginal score}
\label{apx:derivation_marginal}

We derive the marginal score in \Cref{eq:marginal} from the definition $\marg{z_j}{S} \coloneq \Dhatin{S \cup \{z_j\}} - \Dhatin{S}$ in \Cref{sec:method-selection}. The derivation amounts to substituting the estimator $\Dhatin{\cdot}$ into the definition and expanding the resulting quadratic form using the linearity of the aggregate parameter shift.

Recall from \Cref{eq:second-order-influence-in} that the addition-setting estimator is
\begin{equation}
\label{eq:apx-practical-estimator-in}
\Dhatin{S} \;\coloneq\; -\frac{1}{N}\,\grad \target(\thhat)^\top \ushift{S} \;+\; \frac{1}{2 N^2}\,\ushift{S}^\top \Htgt\, \ushift{S},
\end{equation}
and that the aggregate approximated shift over a subset $T$ is $\ushift{T} \coloneq \sum_{z_i \in T} \ushift{i}$. For $T = S \cup \{z_j\}$ with $z_j \notin S$, the linearity of the sum gives
\begin{equation}
\label{eq:apx-aggregate-shift-decomposition}
\ushift{S \cup \{z_j\}} \;=\; \ushift{S} + \ushift{j}.
\end{equation}

Substituting \Cref{eq:apx-aggregate-shift-decomposition} into the linear term of \Cref{eq:apx-practical-estimator-in} evaluated at $S \cup \{z_j\}$ gives
\begin{equation}
\label{eq:apx-marginal-linear}
-\frac{1}{N}\,\grad \target(\thhat)^\top \ushift{S \cup \{z_j\}}
\;=\;
-\frac{1}{N}\,\grad \target(\thhat)^\top \ushift{S}
\;-\;
\frac{1}{N}\,\grad \target(\thhat)^\top \ushift{j}.
\end{equation}
The first term on the right cancels with the corresponding linear term in $\Dhatin{S}$, leaving only the contribution of $z_j$.

Substituting \Cref{eq:apx-aggregate-shift-decomposition} into the quadratic term and expanding gives
\begin{align}
\frac{1}{2 N^2}\,\ushift{S \cup \{z_j\}}^\top \Htgt\, \ushift{S \cup \{z_j\}}
&\;=\;
\frac{1}{2 N^2}\, (\ushift{S} + \ushift{j})^\top \Htgt\, (\ushift{S} + \ushift{j}) \notag \\
&\;=\;
\frac{1}{2 N^2}\, \ushift{S}^\top \Htgt\, \ushift{S}
\;+\;
\frac{1}{N^2}\, \ushift{S}^\top \Htgt\, \ushift{j}
\;+\;
\frac{1}{2 N^2}\, \ushift{j}^\top \Htgt\, \ushift{j},
\label{eq:apx-marginal-quadratic}
\end{align}
where the cross terms $\ushift{S}^\top \Htgt\, \ushift{j}$ and $\ushift{j}^\top \Htgt\, \ushift{S}$ combine into a single term with coefficient $1/N^2$ because $\Htgt$ is symmetric, which gives $\ushift{j}^\top \Htgt\, \ushift{S} = \ushift{S}^\top \Htgt\, \ushift{j}$. The first term on the right of \Cref{eq:apx-marginal-quadratic} is the quadratic term in $\Dhatin{S}$ itself and cancels in the difference $\Dhatin{S \cup \{z_j\}} - \Dhatin{S}$.

Combining the surviving contributions from \Cref{eq:apx-marginal-linear,eq:apx-marginal-quadratic} yields
\begin{equation}
\label{eq:apx-marginal-final}
\marg{z_j}{S}
\;\approx\;
-\frac{1}{N}\,\grad \target(\thhat)^\top \ushift{j}
\;+\;
\frac{1}{N^2}\, \ushift{S}^\top \Htgt\, \ushift{j}
\;+\;
\frac{1}{2 N^2}\, \ushift{j}^\top \Htgt\, \ushift{j},
\end{equation}
which is exactly \Cref{eq:marginal} in the main text. The three surviving terms admit the interpretation given in \Cref{sec:method-selection}: the first term is the standard first-order influence of the candidate $z_j$, the second term measures the curvature-driven interaction between $z_j$ and the already-selected subset $S$, and the third term is the candidate's self-contribution to the curvature correction.

This decomposition aligns with the self versus cross distinction developed in \Cref{apx:derivation_decompositions}. The cross term $\frac{1}{N^2}\, \ushift{S}^\top \Htgt\, \ushift{j}$ in \Cref{eq:apx-marginal-final} is precisely the contribution of $z_j$ to the cross part of the pairwise decomposition, while the self term $\frac{1}{2 N^2}\, \ushift{j}^\top \Htgt\, \ushift{j}$ is its contribution to the diagonal part. The greedy procedure in \Cref{alg:greedy} therefore exploits the same self versus cross structure to update each candidate's marginal utility based on its interaction with previously selected examples, without recomputing the full quadratic form at each iteration.

%% file: tex/999_appendix/D_Hessian_approximations.tex
\section{Scalable Approximations: EK-FAC and Low-Rank Gradient Projection}
\label{apx:hessian_approximation}

The estimators $\Dhatout{S}$ and $\Dhatin{S}$ defined in \Cref{eq:second-order-influence} and \Cref{eq:second-order-influence-in} require applying $H^{-1}$ to per-example gradients. Since $H$ has $p^{2}$ entries in $p$ parameters and is moreover not guaranteed to be positive definite away from a minimum, this is intractable for the models considered in our experiments. We therefore approximate $H$ by the damped Gauss--Newton matrix $G + \lambda I$, where $G$ is the Gauss--Newton Hessian of the training loss at $\hat{\theta}$ and $\lambda > 0$ is a damping constant. This appendix collects the three approximations that make our estimator tractable at scale: the Gauss--Newton approximation itself, its block-diagonal EK-FAC factorization, and a low-rank gradient projection for large language models. The damping constant $\lambda$ is treated as a hyperparameter and its selection is described in \Cref{apx:experiments}.

\subsection{Damped Gauss--Newton approximation}
\label{app:ggn}

Following standard practice in influence-function analysis \citep{bae2022if, grosse2023studying}, we approximate the training Hessian by the Gauss--Newton Hessian $G$ of the training loss at $\hat{\theta}$, damped by $\lambda I$ to ensure positive definiteness.
The Gauss--Newton Hessian is positive semidefinite by construction and discards the indefinite residual term that arises in the exact Hessian, making it a natural surrogate when the goal is to invert a curvature matrix.

The losses considered in this paper are all negative log-likelihoods of exponential-family distributions: cross-entropy with softmax outputs for classification, squared error with a Gaussian likelihood for regression, and token-level negative log-likelihood for instruction tuning.
For such losses, the Gauss--Newton Hessian coincides with the Fisher information matrix taken under the model's output distribution \citep{martens2020new}.
This equivalence is what licenses the use of curvature approximations developed for the Fisher information matrix, and it is the same identification adopted by \citet{grosse2023studying} for influence-function analysis of transformer language models.
We accordingly state our approximations in terms of $G$ throughout, with the understanding that $G$ may equivalently be read as the model-distribution Fisher information.

\subsection{Block-diagonal structure and EK-FAC}
\label{app:ekfac}

Even with the Gauss--Newton substitution, $G$ remains a $p \times p$ matrix and cannot be inverted directly at the parameter counts considered in our experiments.
We therefore further approximate $G$ as block-diagonal across layers, treating cross-layer curvature as zero.
This reduces the inversion to one block per layer.

Within each block, we apply the Eigenvalue-Corrected Kronecker-Factored Approximate Curvature (EK-FAC) approximation \citep{george2018fast, grosse2023studying}.
EK-FAC builds on K-FAC \citep{martens2015optimizing}, which factorizes the per-layer Gauss--Newton block as a Kronecker product of two small matrices: the second-moment matrix of the layer's input activations and the second-moment matrix of the gradient with respect to the layer's pre-activations.
EK-FAC retains the eigenbasis defined by these Kronecker factors but replaces the implied Kronecker-structured eigenvalues with the exact diagonal of $G$ in that eigenbasis, estimated from training data.
The resulting approximation is provably at least as accurate as K-FAC under the Frobenius norm \citep{george2018fast} while preserving the same favorable scaling: each per-layer block can be inverted, and inverse-vector products evaluated, at a cost dominated by the layer's activation and pre-activation dimensions rather than its full parameter count.
In practice, the per-example shifts $u_i$ required by \Cref{eq:second-order-influence} and \Cref{eq:second-order-influence-in} are computed as $(G + \lambda I)^{-1} \nabla_{\theta} \ell(z_{i}, \hat{\theta})$ by applying this block-wise EK-FAC inverse to per-example gradients. We refer the reader to \citet{george2018fast} for the original derivation of EK-FAC and to \citet{grosse2023studying} for a detailed treatment of its use in influence-function analysis.

\subsection{Low-rank gradient projection for large language models}
\label{app:logra}

At the scale of Llama-3.1-8B, the per-layer EK-FAC blocks remain large enough that materializing and storing one projected gradient $u_{i}$ per training example, as required by the greedy procedure in \Cref{alg:greedy}, is itself the dominant cost.
To address this, we adopt the low-rank gradient projection of LoGra \citep{choe2024your}, which introduces a shared low-dimensional subspace and projects all per-example gradients into that subspace before any subsequent inner product or curvature operation is performed.
The projection is applied consistently to gradients and to the target-side quantities entering \Cref{alg:greedy}, so that all inner products in the greedy loop are evaluated in the projected dimension $d$ rather than in the full parameter dimension $p$.

This projection is what reduces the per-iteration cost of \Cref{alg:greedy} to $O(P)$ inner products in dimension $d$, yielding the overall greedy-loop complexity of $O(KPd)$ stated in \Cref{sec:method-selection}.
We use the projection dimension and hyperparameters reported in \Cref{apx:experiments}, and refer the reader to \citet{choe2024your} for the construction of the projection operator and a full treatment of the resulting estimator.

%% file: tex/999_appendix/E_experimental_settings.tex
\section{Experimental Details}
\label{apx:experiments}

\subsection{Small-scale Attribution Experiments}

The attribution experiments in \Cref{sec:exp_attribution} consider six 
dataset--model pairs: logistic regression and MLP on MNIST, MLP on 
FashionMNIST, MLP on Concrete and Parkinsons, and ResNet-9 on CIFAR-10. 
All datasets use their standard train/test splits.

\paragraph{Models.}
LR uses $\ell_2$ regularization with strength $0.01$. The MLP has two 
hidden layers of width $128$ and $64$ with ReLU activations. ResNet-9 
follows the standard DAWNBench architecture~\citep{coleman2017dawnbench}, 
trained from scratch.

\paragraph{Training.}
Training configurations are summarized in \Cref{tab:model_config} and 
follow standard settings for each model.
\input{fig/src/tab_apx_small_scale_config}

\paragraph{Group construction.}
Each group consists of an anchor sampled uniformly at random from the 
training set, together with its $|S|-1$ nearest neighbors in softmax 
output space (under $L^{2}$ distance). We set $|S| = 400$, yielding 50 
subsets per dataset--model pair, except for Concrete where we set $|S| 
= 100$ due to its small training set. For regression tasks, the 
prediction itself replaces the softmax output.

\paragraph{Ground truth.}
For each subset $S$, we retrain the model from scratch on $\mathcal{D} 
\setminus S$ and report the mean change in held-out test loss.

\paragraph{Influence computation.}
For LR, the Hessian is computed and inverted exactly. For MLPs and 
ResNet-9, we use the damped Gauss--Newton approximation of the training 
Hessian, $G + \lambda I$, and approximate its inverse via EK-FAC. The 
damping $\lambda$ is set to $10^{-2}$ for all influence-based methods. 
The target Hessian $H_f$ is approximated as block-diagonal, with 
$H_f$-vector products computed directly without inversion.

\paragraph{Baselines.}
\emph{F} is the standard first-order influence function~\citep{koh2017understanding}. 
\emph{F+B} applies the second-order response-function correction of 
\citet{basu2020second}, computed under the same curvature approximation 
as our method. \emph{F+I+B} combines our interaction term with the Basu 
correction under identical approximations. For TRAK~\citep{park2023trak}, 
we use projection dimension $2048$ and ridge $0.01$. For 
TracIn~\citep{pruthi2020estimating}, we use $10$ evenly spaced checkpoints. 
For both TRAK and TracIn, group scores are obtained by summing individual 
attributions.

\subsection{Small-Scale Selection Experiments}
\label{apx:exp_small_scale_selection}

The selection experiments in \Cref{sec:exp_validation_small_scale} use a two-hidden-layer MLP ($784 \rightarrow 128 \rightarrow 64 \rightarrow 10$) with ReLU activations and dropout rate $0.1$. For each dataset, a candidate pool $\Dpool$ of $5{,}000$ training examples is drawn by a fixed permutation of the original training set, and the reference parameter $\thhat$ is obtained by training the MLP on this pool for $200$ epochs with vanilla SGD (learning rate $10^{-2}$, batch size $64$, weight decay $3\!\times\!10^{-2}$). The Hessian-inverse $\Htr^{-1}$ is approximated via EK-FAC with the GGN Fisher and damping $\lambda = 5\!\times\!10^{-1}$. Each selection picks $K\!\in\!\{500, 1{,}000, \ldots, 5{,}000\}$ examples in increments of $500$ from the same pool $\Dpool$, and each selected subset is retrained from scratch with the same training configuration. Held-out test loss is averaged over five random seeds and reported; the class-entropy panel reports the Shannon entropy of the empirical class distribution among the $K$ selected examples.

\input{fig/src/tab_apx_selection_cost}
\input{fig/src/tab_apx_our_cost}

\subsection{LLM Data Selection Experiments}
\label{apx:exp_llm_selection}

We fine-tune Llama-3.1-8B on subsets selected from the LESS pool of approximately $270$K instruction-tuning examples.

\paragraph{Warmup and selection.}
Following~\citet{xia2024less}, a warmup model $\hat{\theta}$ is obtained by fine-tuning the base model on a random subset of size $13{,}534$ using LoRA (rank $8$, scaling $16$, applied to \texttt{q\_proj}, \texttt{k\_proj}, \texttt{v\_proj}, \texttt{o\_proj}, \texttt{gate\_proj}, \texttt{up\_proj}, and \texttt{down\_proj}), AdamW with peak learning rate $10^{-4}$, cosine schedule with warmup ratio $0.03$, effective batch size $128$, sequence length $2048$, in bf16, for four epochs.
At $\hat{\theta}$, we compute per-example projected gradients $\ushift{j}$ and curvature terms $\Htgt \ushift{j}$ using LoGra~\citep{choe2024your} with projection dimension $16$, and apply \Cref{alg:greedy} to select $K = 13{,}534$ examples (5\% of the pool) independently for each target task.
The target function $f$ is the instruction-masked likelihood, evaluated on the official task validation set when available and on a held-out training split otherwise.
%The selected indices for all seven target tasks are released alongside our code, enabling reproduction of the fine-tuning results without rerunning the selection pipeline.

\paragraph{Final fine-tuning.}
Selected subsets are used to fine-tune the original pretrained Llama-3.1-8B checkpoint (rather than the warmup model) under the same LoRA configuration as warmup: AdamW with peak learning rate $10^{-4}$, cosine schedule with warmup ratio $0.03$, effective batch size $128$, sequence length $2048$, and bf16 precision.
We train for four epochs on GSM8K, whose loss does not converge within a single epoch, and one epoch on the remaining tasks.

\paragraph{Evaluation.}
We evaluate on GSM8K, AQuA, ARC-Easy, HellaSwag, PIQA, ECQA, and SQuAD using the Tulu chat template~\citep{wang2023far}.
All tasks use 0-shot prompting.
We report exact match for GSM8K, F1 for SQuAD, and accuracy for the multiple-choice tasks, with the latter scored by length-normalized log-likelihood.
Results are averaged over five random seeds.

\paragraph{Baselines.}
All methods share the same candidate pool, selection budget, and final fine-tuning protocol; influence-based methods additionally share the same warmup checkpoint.
Additive IF uses our warmup checkpoint and projected gradients but selects the top-$K$ examples by first-order influence, omitting the interaction term.
LESS~\citep{xia2024less} uses Adam-adjusted gradient cosine similarity with projection dimension $8192$ on the same warmup checkpoint.
RDS+~\citep{ivisondata2025} uses position-weighted mean-pooled hidden states from the warmup checkpoint with cosine similarity.
NV-Embed~\citep{lee2024nv} uses embeddings from \texttt{nvidia/NV-Embed-v2}.
Random uniformly subsamples from the pool.

\subsection{Compute Resources}
\label{apx:compute}

All small-scale experiments were conducted on a single NVIDIA RTX A6000 (48\,GB), and all LLM experiments on $8 \times$ NVIDIA RTX A6000 (48\,GB).

\Cref{tab:selection_cost} reports the GPU-hours required by each selection method on the LESS pool of approximately $270$K instruction-tuning examples, summed over all devices used.
Warmup fine-tunes a LoRA adapter on a random subset of size $13{,}534$ for four epochs.
Additive IF and our method share the same warmup checkpoint and LoGra-projected gradients, so their selection costs differ only in the greedy step, which contributes approximately $50$ seconds per target.

\Cref{tab:our_pipeline_cost} decomposes the selection cost of our pipeline into its individual components.
Warmup fine-tuning and pool gradient extraction are shared across all seven targets, while the selection step is paid once per target.
The selection step itself further decomposes into four sub-steps: a one-time covariance-state merge that is shared across targets, validation-set gradient extraction that scales per target, first-order influence and target-curvature inner products, and the greedy selection loop.
These four sub-steps together account for the $0.07$ GPU-hours per target reported in \Cref{tab:our_pipeline_cost}.

%% file: fig/src/tab_apx_small_scale_config.tex
\begin{table}[h]
\centering
\small
\caption{Training configurations for small-scale models. The cosine schedule decays the learning rate to zero by the end of training.}
\label{tab:model_config}
\begin{tabular}{lccccccc}
\toprule
Model & Optimizer & Momentum & LR & LR Schedule & Weight Decay & Batch Size & Epochs \\
\midrule
LR        & SGD & 0.0 & 0.01 & constant & 0.01 & 64 & 200 \\
MLP       & SGD & 0.0 & 0.01 & constant & 0.01 & 64 & 200 \\
ResNet-9  & SGD & 0.9 & 0.01 & cosine   & 0.01 & 64 & 50  \\
\bottomrule
\end{tabular}
\end{table}

%% file: fig/src/tab_apx_selection_cost.tex
\begin{table}[t]
\centering
\small
\caption{Compute cost of selection methods on the LESS pool with Llama-3.1-8B, in GPU-hours.}
\label{tab:selection_cost}
\begin{tabular}{lccc}
\toprule
Method & Warmup & Selection & Total \\
\midrule
Additive IF                     & $9.2$ & $22$  & $31$  \\
LESS~\citep{xia2024less}        & $9.2$ & $306$ & $315$ \\
RDS+~\citep{ivisondata2025}     & --    & $23$  & $23$  \\
NV-Embed~\citep{lee2024nv}      & --    & $109$ & $109$ \\
\textbf{Ours}                   & $9.2$ & $22$  & $31$  \\
\bottomrule
\end{tabular}
\end{table}

%% file: fig/src/tab_apx_our_cost.tex
\begin{table}[t]
\centering
\small
\caption{Per-component GPU-hours of our selection pipeline on the LESS pool with Llama-3.1-8B.}
\label{tab:our_pipeline_cost}
\begin{tabular}{lcc}
\toprule
Component & GPU-hours & Scope \\
\midrule
Warmup fine-tuning                & $9.2$  & shared \\
Pool gradient extraction (LoGra)  & $22$   & shared \\
Selection step                    & $0.07$ & per target \\
\bottomrule
\end{tabular}
\end{table}

%% file: tex/999_appendix/F_additional_experiments.tex
\section{Additional Experiments}
\label{apx:additional_exp}
\input{fig/src/fig_apx_additional_exp}
 
In this section, we present two additional experiments on the two-layer MLP that complement 
the small-scale attribution accuracy results of \Cref{sec:exp_attribution} 
and the small-scale selection quality results of \Cref{sec:exp_validation_small_scale}. 
The first examines the sensitivity of each method to the damping 
hyperparameter $\lambda$ used in the EK-FAC, and the second extends the 
class entropy analysis from MNIST to FashionMNIST.
The results are summarized in \Cref{fig:apx_additional_exp}.

\paragraph{Setup.}
Both experiments inherit the protocols of their counterparts in the main text, modified as follows.
For the damping sensitivity analysis, we follow the attribution setup of \Cref{sec:exp_attribution} but vary $\lambda \in \{0.005, 0.01, 0.02, 0.05, 0.1\}$ instead of fixing $\lambda = 10^{-2}$.
All other elements of the protocol, including group construction, ground-truth retraining, and the influence-function methods (\emph{F}, \emph{F+B}, \emph{F+I+B}, Ours), are unchanged.
For the class entropy analysis, we run the selection procedure of \Cref{sec:exp_validation_small_scale} on FashionMNIST instead of MNIST, keeping all other choices unchanged.
 
\paragraph{Damping sensitivity.}
The left and middle panels of \Cref{fig:apx_additional_exp} report the Spearman correlation as a function of the damping $\lambda$ on MNIST and FashionMNIST.
Ours and F+I+B remain nearly constant across the tested range on both 
datasets, while F and F+B are somewhat less stable, with F+B showing 
the most pronounced decline at larger $\lambda$ on MNIST.
%Since $\lambda$ enters the EK-FAC inverse and thereby affects the per-example parameter shifts $u_i$, this stability suggests that small differences in the parameter-shift estimates do not translate into a visible loss of attribution accuracy in the settings we evaluate.
 
\paragraph{Class entropy on FashionMNIST.}
The right panel of \Cref{fig:apx_additional_exp} reports the class entropy of the subsets selected by each method on FashionMNIST.
% The pattern matches the MNIST result reported in \Cref{sec:exp_validation_small_scale} (Figure 3, right): at small budgets the first-order influence function (F) and its Basu-corrected variant (F+B) collapse onto a few classes, while our method matches the entropy of random selection across all selection sizes.
The pattern matches the MNIST result reported in \Cref{fig:small_selection}: at small budgets, F and F+B collapse onto a few classes and 
F+I+B shows a milder but still noticeable drop, while our method 
matches the entropy of random selection across all selection sizes.
This confirms that the diversity-promoting behavior of the interaction term is not specific to MNIST and carries over to FashionMNIST.

%% file: fig/src/fig_apx_additional_exp.tex
\begin{figure}[t]
    \centering
    \includegraphics[width=\linewidth]{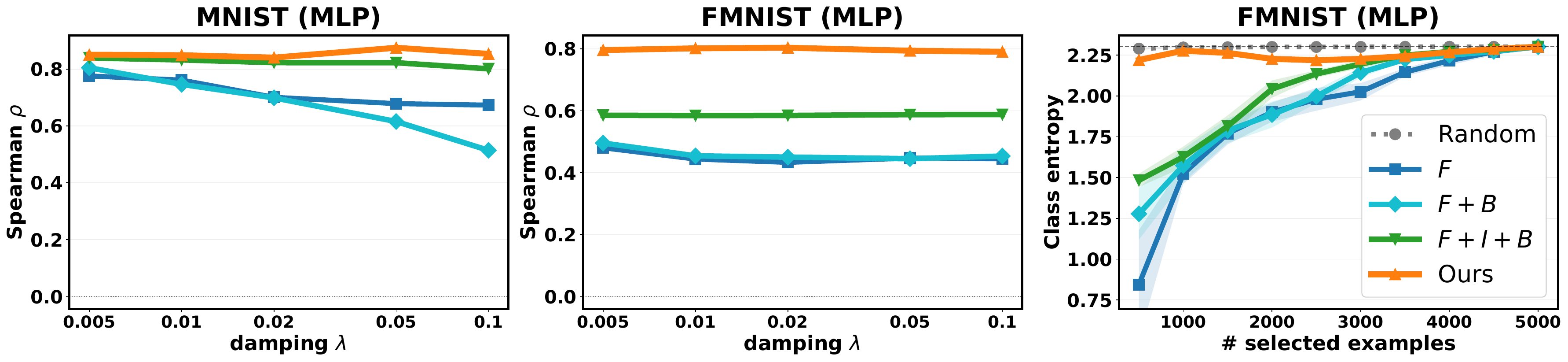}
    \caption{Sensitivity analysis of the two-layer MLP over the damping hyperparameter on MNIST (left) and FashionMNIST (middle), and class entropy on FashionMNIST (right).}
    \label{fig:apx_additional_exp}
\end{figure}